\pdfoutput=1

\documentclass[11pt]{article}

\usepackage[final]{acl}

\usepackage{times}
\usepackage{latexsym}

\usepackage{booktabs}
\usepackage{rotating}
\usepackage{amssymb,amsmath}
\usepackage[dvipsnames]{xcolor}

\newcommand{\red}[1]{\textcolor{red}{#1}}
\newcommand{\green}[1]{\textcolor{Green}{#1}}

\usepackage[T1]{fontenc}

\usepackage[utf8]{inputenc}

\usepackage{microtype}

\usepackage{inconsolata}

\usepackage{graphicx}

\usepackage{listings}

\title{Extracting Conceptual Spaces from LLMs Using Prototype Embeddings}

\author{Nitesh Kumar \and Usashi Chatterjee \and Steven Schockaert\\
  Cardiff NLP, School of Computer Science and Informatics \\
  Cardiff University, United Kingdom \\
  \texttt{\{kumarn8,chatterjeeu,schockaerts1\}@cardiff.ac.uk} \\}

\begin{document}

\maketitle

\begin{abstract}
Conceptual spaces represent entities and concepts using cognitively meaningful dimensions, typically referring to perceptual features. Such representations are widely used in cognitive science and have the potential to serve as a cornerstone for explainable AI. Unfortunately, they have proven notoriously difficult to learn, although recent LLMs appear to capture the required perceptual features to a remarkable extent. Nonetheless, practical methods for extracting the corresponding conceptual spaces are currently still lacking. While various methods exist for extracting embeddings from LLMs, extracting conceptual spaces also requires us to encode the underlying features. In this paper, we propose a strategy in which features (e.g.\ \emph{sweetness}) are encoded by embedding the description of a corresponding prototype (e.g.\ \emph{a very sweet food}). To improve this strategy, we fine-tune the LLM to align the prototype embeddings with the corresponding conceptual space dimensions. Our empirical analysis finds this approach to be highly effective.
\end{abstract}

\section{Introduction}

Conceptual spaces \cite{DBLP:books/daglib/0006106} are geometric representations of meaning, in which concrete entities are represented as vectors. Different from word embeddings in NLP, the dimensions of a conceptual space (typically) correspond to perceptual features. For instance, in a colour space, entities would be represented using three dimensions, corresponding to their hue, saturation and intensity. Conceptual spaces are used in cognitive science as theoretical models to explain phenomena such as analogy \cite{DBLP:journals/jetai/OstaVelezG24}, non-monotonic reasoning \cite{DBLP:journals/jolli/Osta-VelezG22} and concept learning \cite{DBLP:journals/mima/Douven23}. Within AI, the use of conceptual spaces has been advocated as an interface between neural and symbolic representations \cite{DBLP:journals/ai/AisbettG01}. As such, they can play an important role in explainable AI, for instance to enable interpretable classifiers \cite{DBLP:journals/ai/DerracS15,DBLP:journals/jair/BanaeeSL18,DBLP:journals/corr/abs-2502-13632} and computational creativity \cite{DBLP:journals/jagi/McGregorAPW15}. In practice, however, these applications have been hampered by the difficulty in learning conceptual spaces. Within cognitive science, most work has relied on spaces that are learned from human similarity judgments, for instance to study perception of colour \cite{DBLP:journals/cogsci/DouvenWJD17}, music \cite{DBLP:journals/mima/ForthWM10}, taste \cite{Paradis2015} or smell \cite{DBLP:journals/cogsci/JraissatiD21}. Clearly, however, such a solution is not scalable enough for explainable AI. 

A natural alternative is to try to construct conceptual spaces using NLP models, such as word embeddings or Large Language Models (LLMs). In fact, even within cognitive science, researchers have looked at NLP models as a promising route to obtain conceptual spaces in a cheaper way \cite{DBLP:journals/mima/MoullecD25}. Starting from a pre-trained embedding space, it is often indeed possible to identify directions within that space that capture meaningful ordinal properties \cite{gupta-etal-2015-distributional,DBLP:journals/ai/DerracS15,gari-soler-apidianaki-2020-bert,grand2022semantic,erk-apidianaki-2024-adjusting}. However, modelling \emph{perceptual} features with traditional models has proven more challenging. This is intuitively due to the fact that many perceptual features are only rarely stated in text. For instance, \citet{paik-etal-2021-world} highlighted how language models struggle with predicting colours, due to a divergence between the typical colour of an object and the distribution of co-occurring colour terms (e.g.\ the phrase ``green banana'' being more common than ``yellow banana'' in text). However, recent LLMs have proven more capable at modelling perceptual features, where promising results have been reported for colour \cite{abdou-etal-2021-language,liu-etal-2022-ever,DBLP:conf/iclr/PatelP22,marjieh2024large}, taste \cite{kumar-etal-2024-ranking,marjieh2024large}, touch \cite{DBLP:journals/corr/abs-2406-06587}, smell \cite{DBLP:journals/corr/abs-2411-06950} and sound \cite{marjieh2024large}, among others.

One problem that is not addressed by these works is how to \emph{extract} conceptual spaces from LLMs. For instance, 
\citet{kumar-etal-2024-ranking} prompt LLMs to make pairwise judgments (e.g.\ which is sweeter, banana or cucumber?), which only allows us to \emph{rank} the entities along some conceptual space dimensions, without capturing how much the entities differ. Using pairwise comparisons is also intractable when dealing with thousands of entities. \citet{marjieh2024large} use LLMs to make pairwise similarity judgments, which is again too inefficient for constructing conceptual spaces at scale. 

In a wider context, the problem of learning embeddings of text fragments using LLMs is well-studied \cite{reimers-gurevych-2019-sentence,gao-etal-2021-simcse,liu-etal-2021-fast,wang-etal-2024-improving-text,DBLP:journals/corr/abs-2404-05961,DBLP:journals/corr/abs-2405-17428}. We therefore consider the following research question: is it possible to extract conceptual spaces directly from LLM-generated embeddings? Entity embeddings can straightforwardly be obtained using standard techniques. However, we also need to model the perceptual features. For instance, given an embedding $\textit{emb}(\textrm{``banana''})$ of the word banana, how do we determine its level of sweetness? As already mentioned, previous work has shown that many features of interest can be modelled as directions in pre-trained embeddings. It is thus natural to assume that there exists a vector $\mathbf{v}_{\scriptsize\textit{sweet}}$ such that $\textit{emb}(\textrm{``banana''})\cdot \mathbf{v}_{\scriptsize\textit{sweet}}$ reflects the degree of sweetness of a banana. One possibility is to estimate this vector $\mathbf{v}_{\scriptsize\textit{sweet}}$ from labelled examples, but such data is not readily available for most domains. Another possibility is to estimate the vector from seed words, i.e.\ examples of entities at both extremes of the ranking, but such directions can be unreliable, being highly sensitive to choice of seeds \cite{antoniak-mimno-2021-bad,erk-apidianaki-2024-adjusting}.

\begin{figure}
\includegraphics[trim={90pt 0 110pt 0},clip,width=\columnwidth]{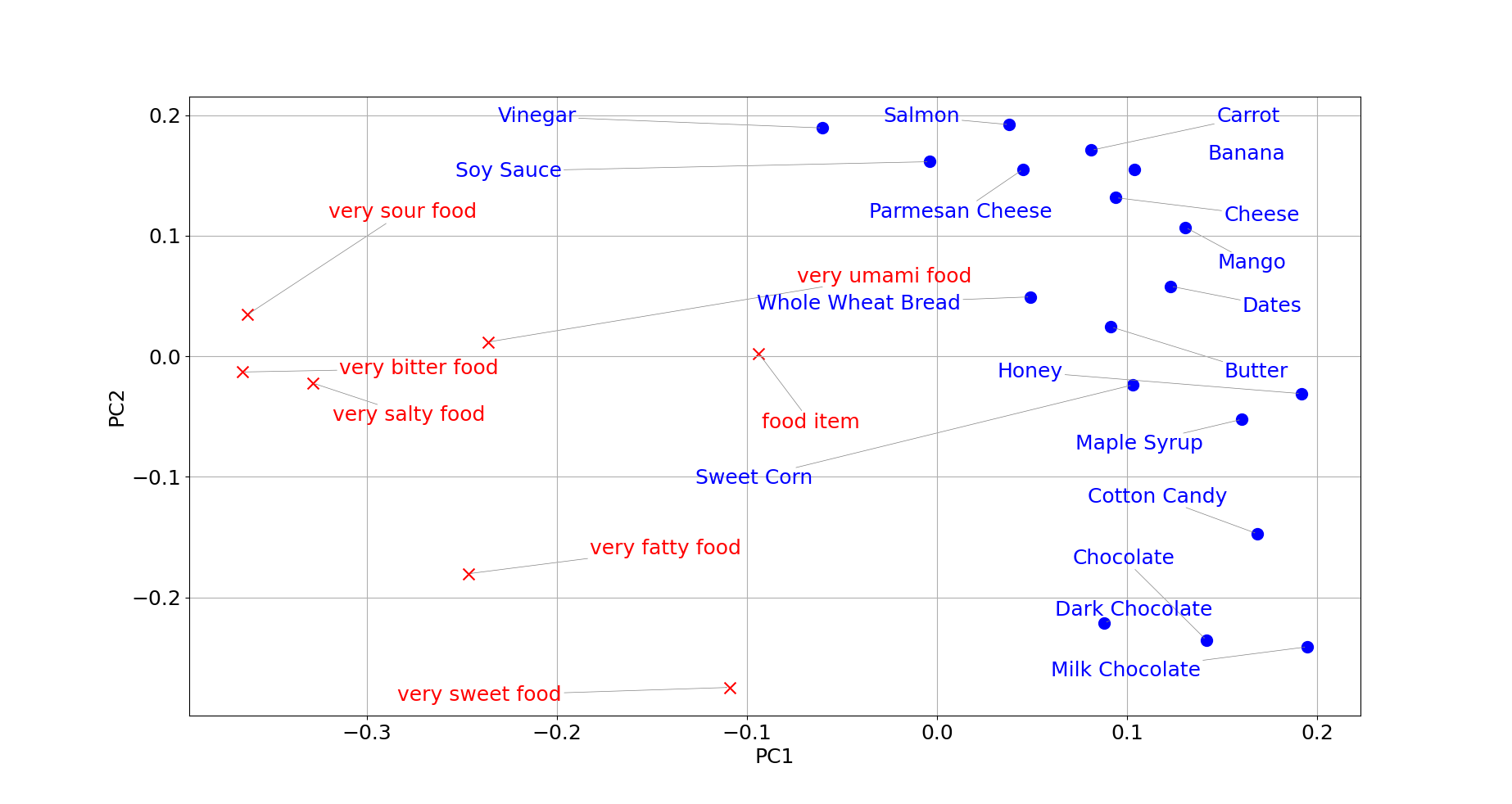}
\includegraphics[trim={90pt 0 110pt 0},clip,width=\columnwidth]{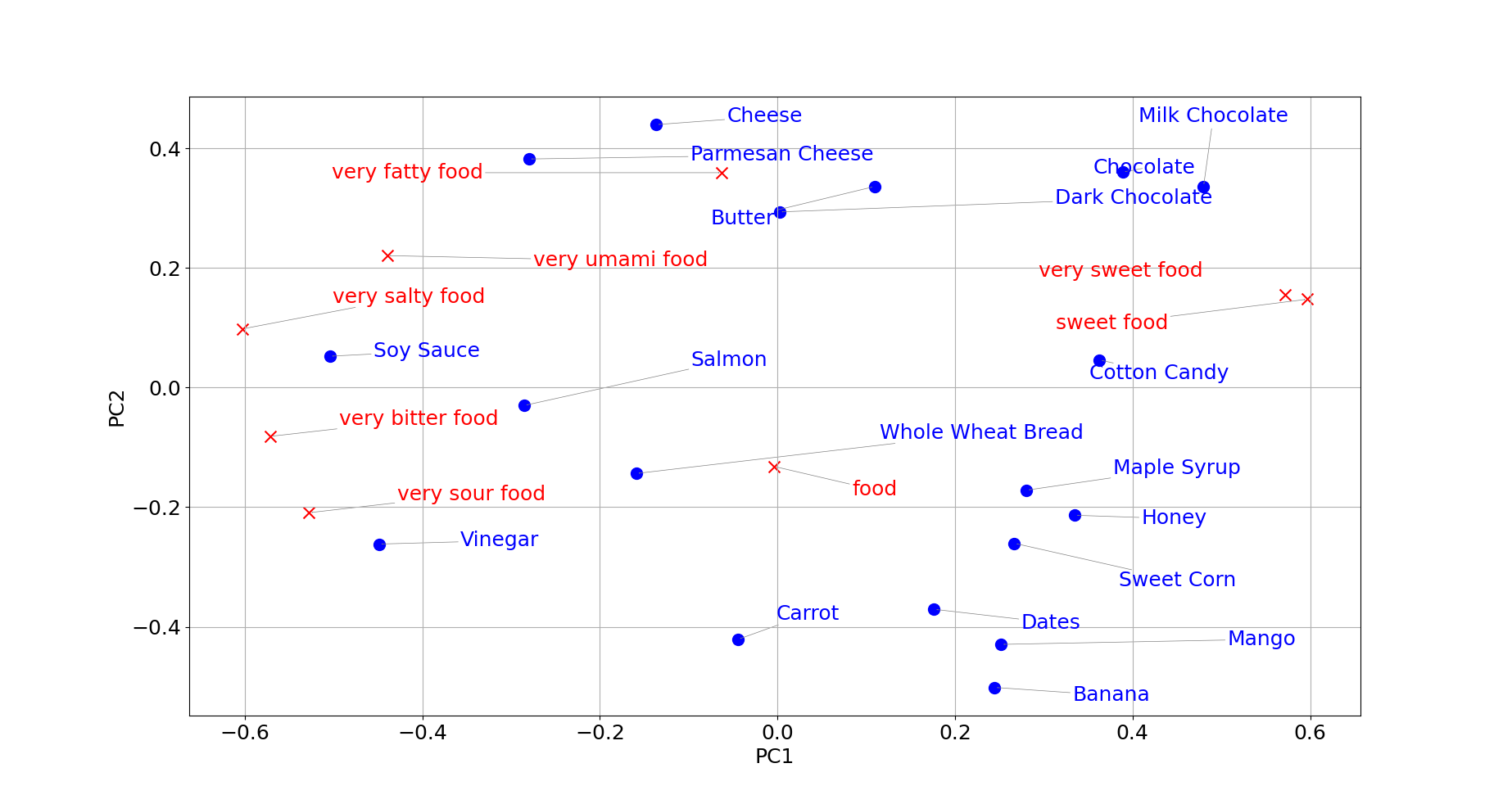}
\caption{Embeddings of entities and prototypes in pre-trained LLM embedding models (top) and after fine-tuning (bottom), showing the first two principal components. 
\label{figFinetuningEffect}}
\end{figure}

In this paper, we consider a simple alternative, which is to estimate the vector $\mathbf{v}_{f}$ encoding some feature $f$ as the description of a generic prototype. For instance, $\mathbf{v}_{\scriptsize\textit{sweet}}$ could be modelled as $\textit{emb}(\textrm{``a very sweet food''})$. Unfortunately, with pre-trained LLM embedding models, the performance of this approach is sub-optimal, as the embedding of such a generic prototype description lies in a different subspace than the entities themselves (see Figure \ref{figFinetuningEffect}). We therefore propose a fine-tuning strategy, which encourages the embeddings of such descriptions to be aligned with the embeddings of the corresponding entities. We find that a small training set, synthetically generated using GPT-4o, is sufficient to achieve state-of-the-art results.

\section{Related Work}
The problem of learning entity embeddings using language models has received considerable attention, especially for bidirectional models of the BERT family \cite{devlin-etal-2019-bert}. For instance, a number of authors have proposed to represent entities by averaging the contextualised embeddings of their mentions in a corpus, using pre-trained \cite{ethayarajh-2019-contextual,bommasani-etal-2020-interpreting,vulic-etal-2020-probing,liu-etal-2021-mirrorwic} or fine-tuned \cite{DBLP:conf/sigir/LiKBS23} language models. However, approaches that directly extract embeddings based on the name of an entity have also been studied \cite{vulic-etal-2021-lexfit,liu-etal-2021-fast,gajbhiye-etal-2022-modelling}. Most relevant to our work, several authors have focused on predicting semantic and commonsense properties of concepts from their embeddings \cite{gajbhiye-etal-2022-modelling,DBLP:conf/sigir/LiKBS23,DBLP:journals/nle/RosenfeldE23}. For instance, \citet{chatterjee-etal-2023-cabbage} evaluated a BERT encoder that was fine-tuned to predict commonsense properties on the task of predicting taste dimensions such as sweetness, showing that their encoder was able to match the performance of GPT-3. \citet{kumar-etal-2024-ranking} showed that a fine-tuned Llama 3 model is able to outperform BERT encoders. 
In this paper, we build on these results, aiming to extract embeddings from models such as Llama3, rather than using them for making pairwise judgments.

Compared to encoder-only models such as BERT, it is somewhat less straightforward to use decoder-only LLMs for embedding text. However, in recent years, several successful strategies have been proposed for fine-tuning LLMs to become general-purpose text embedding models \cite{wang-etal-2024-improving-text,DBLP:journals/corr/abs-2404-05961,DBLP:journals/corr/abs-2405-17428}, with the Massive Text Embedding Benchmark \cite{muennighoff-etal-2023-mteb} serving as a key driver. However, the focus of this benchmark is on sentence and paragraph level tasks, and little is currently known about the quality of LLM embedding models when it comes to representing entities. Our analysis in this paper partially addresses this gap, by comparing the quality of the conceptual space representations that are obtained by several recent models. LLMs can also be used to predict embeddings without fine-tuning. \citet{jiang-etal-2024-scaling} suggested an Explicit One word Limitation (EOL) prompt, of the following form, for this purpose: \emph{``This sentence: [text] means in one word:''.} We will also rely on prompts with this one-word limitation.


\section{Methodology}\label{secMethodology}
\paragraph{Problem Formulation}
Let $\textit{emb}$ be an LLM-based embedding model, where we write $\textit{emb}(x)\in\mathbb{R}^n$ for the encoding of a phrase $x$. Let us furthermore assume that a set of entities $\mathcal{E}$ is given which all belong to some natural category. For instance, the entities in $\mathcal{E}$ could represent different types of food (e.g.\ banana, roast chicken, cake). For an entity $e$, we write $\gamma(e)$ for the verbalization of that entity, i.e.\ $\gamma(e)$ is a phrase that describes $e$. The entity $e$ can then be represented by its embedding $\textit{emb}(\gamma(e))$. We are interested in modelling semantic features of the entities based on these embeddings, where our focus is on perceptual features such as the sweetness of a  food item or the intensity of an odour. Let $f$ be some real-valued feature, such that every entity $e\in\mathcal{E}$ has a corresponding feature value $f(e)\in\mathbb{R}$. We want to find an encoding $\tau_f:\mathbb{R}^n\rightarrow\mathbb{R}$ of the feature $f$ such that $\tau_f(\textit{emb}(\gamma(e)))\in \mathbb{R}$ corresponds to the feature value $f(e)$. We want to find the encoding $\tau_f$ without any supervision, other than a verbalization of the feature $f$, hence we cannot expect $\tau_f(\textit{emb}(\gamma(e)))=f(e)$, as there is typically no unique way to measure the degree to which a perceptual feature is satisfied. Instead, we want the \emph{rankings} induced by the functions $f(.)$ and $\tau_f(\textit{emb}(\gamma(.)))$ to be as similar as possible.


\paragraph{Embedding Entities}
We experiment with two types of models: standard LLMs such as Llama-3 and pre-trained LLM-based embeddings models such as E5. The latter models can directly be used to obtain an embedding of $\gamma(e)$. To obtain embeddings with standard LLMs, we use a variant of the EOL trick from \citet{jiang-etal-2024-scaling}. Specifically, we use the following prompt:

\medskip
\noindent\textit{The description of the term `$\gamma(e)$' in one word is}

\medskip
\noindent The embedding $\textit{emb}(\gamma(e))$ is then defined as the \textit{normalized} encoding of the LLM for the last token. To verbalize the entity $e$, we observed that adding the name of the considered category leads to more informative embeddings for most models. For instance, we verbalize the entity \emph{banana} as ``food item banana'' rather than ``banana''. This intuitively helps with resolving some ambiguities (e.g.\ orange as a fruit rather than a colour) and with specializing the embeddings to the domain of interest (e.g.\ strawberry as an odour rather than a food). 

\paragraph{Modelling Features}
A common approach for modelling semantic features based on embeddings is to fit a logistic regression model (or a linear SVM) based on some training data. However, for most perceptual features, such training data is not readily available. Another common approach relies on a few examples of seed words $h_1,...,h_p$ which are known to have a high value for the considered feature, and examples of seed words $l_1,...,l_q$ which are known to have a low value. We can then estimate a vector $\mathbf{v}_f$ that models the considered feature $f$ based on this vectors, e.g.:
\begin{align*}
\mathbf{v}_f = \frac{1}{p}\sum_{i=1}^p \textit{emb}(\gamma(h_i)) - \frac{1}{q}\sum_{i=1}^q \textit{emb}(\gamma(l_i))
\end{align*} 
and $\tau_f(\mathbf{e})= \mathbf{e}\cdot \mathbf{v}_f$.
In principle, $\mathbf{v}_f$ can then be estimated from just two seeds words (i.e.\ $p=q=1$). However, several authors have pointed out that this approach can be unreliable \cite{antoniak-mimno-2021-bad,erk-apidianaki-2024-adjusting}. For instance, if we have \emph{banana} as the only example of a sweet food, then the resulting vector $\mathbf{v}_{\textrm{sweetness}}$ might capture the property of being yellow (in addition to, or instead of sweetness).

We pursue a different strategy, estimating the vector $\mathbf{v}_f$ by embedding a description $\gamma(f)$ of the feature $f$. \citet{gajbhiye-etal-2022-modelling} trained a BERT bi-encoder based on this idea. Specifically, they fine-tuned two different BERT models, one for encoding entities and one for encoding properties, using a large dataset of commonsense properties. With LLMs, this bi-encoder strategy is not practical, as it doubles the memory requirement compared to fine-tuning a single model. We therefore embed entities and features using the same model. However, we still need to ensure that the embeddings of entities and features are aligned, i.e.\ $\textit{emb}(\gamma(e))\cdot \textit{emb}(\gamma(f))$ should reflect the extent to which $e$ has the feature $f$. To this end, we verbalize $f$ as a generic description of a prototypical entity with a high value for the feature $f$. For instance, we can choose:
\begin{align*}
\gamma(\text{sweetness}) = \text{``a very sweet food''}
\end{align*}
However, as illustrated in Figure \ref{figFinetuningEffect}, the embeddings of such generic descriptions are not in the same subspace as those of the entities. We therefore add a fine-tuning step, as we explain next.

\paragraph{Fine-tuning Strategy}
We fine-tune the embedding model $\textit{emb}$ to encourage the encoding of a generic property to be similar to the encoding of entities that have that property. For instance we want $\textit{emb}(\text{``a tall mountain''})$ to be similar to $\textit{emb}(\text{``Mount Everest''})$. To this end, we collected a small dataset using GPT-4o, consisting of information about 123 target properties. For each target property (e.g.\ \emph{long river}), the dataset lists 7 examples of entities which have this property (e.g.\ \emph{Nile}, \emph{Amazon}, \emph{Yangtze}), as well as 4 negative properties, which the entities do not satisfy. Of these negative properties, 3 are closely related to the target property (e.g.\ \emph{short river}) and one is non-sensical for the considered entity type (e.g.\ \emph{small city} when the entities are rivers).\footnote{Appendix \ref{appFinetuningDataset} provides more details about the dataset.} 
We encourage the target property embedding to be close to the centroid of the seven examples and further from the negative properties. Specifically, we fine-tune the LLM by minimizing the following \textit{classification loss}:
\begin{align*}
- \log \frac{ \exp \left( \frac{ \textit{emb}(\gamma(f_0)) \cdot \mathbf{c} }{T} \right) }{ \sum_{k=0}^{4} \exp \left( \frac{ \textit{emb}(\gamma(f_k)) \cdot \mathbf{c} }{T} \right) }
\end{align*} 
where $\mathbf{c}$ is the centroid of entity embeddings, i.e., $\mathbf{c} = \frac{1}{7}\sum_{i=1}^{7}emb(\gamma(e_i))$, $f_0$ is the target property, and $f_1, \dots, f_4$ are negative properties, with $T>0$ a temperature parameter. 
We write $\mathcal{L}_1$ for the average classification loss across all target properties. 

Note that the fine-tuning process explained thus far does not specifically focus on perceptual features, nor on the fact that we use the embeddings for ranking. \citet{kumar-etal-2024-ranking} found that models which were fine-tuned on perceptual features generalized well to other, previously unseen perceptual features. As a secondary fine-tuning objective, we therefore also include the following \textit{ranking loss}: 
\begin{align*}
\sigma\left( -\alpha \cdot y_i \cdot \left[ (\mathbf{e}_{1} - \mathbf{e}_{2}) \cdot emb(\gamma(f)) \right] \right)
\end{align*}
where $y_i \in \{-1, +1\}$ indicates whether $e_1$ should rank above $e_2$ with respect to feature $f$, $\alpha$ is a scaling hyperparameter, $\mathbf{e}_1 = emb(\gamma(e_1))$, $\mathbf{e}_2 = emb(\gamma(e_2))$, and $\sigma$ denotes the sigmoid function. We write $\mathcal{L}_2$ for the average ranking loss across all entity pairs in our training set. The overall loss is then simply given by $\mathcal{L}_1+ \lambda \mathcal{L}_2$, where $\lambda$ is a hyperparameter. 

\section{Datasets}\label{secDatasets}
Following \citet{kumar-etal-2024-ranking}, we evaluate our approach on the following datasets:
\begin{description}
\item[Taste:] a dataset, originally created by \citet{martin2014creation}, describing the taste of 590 food items, in terms of the following quality dimensions: sweetness, sourness, saltiness, bitterness, fattiness and umaminess. This dataset was first used for evaluating LLMs by \citet{chatterjee-etal-2023-cabbage}, who rephrased some of the properties to make the more suitable for prompting. We use their cleaned version of the dataset.
\item[Rocks:] a dataset, originally created by \citet{nosofsky2018toward}, describing the physical appearance of 30 types of rocks, in terms of the following dimensions: lightness of colour, average grain size, roughness, shininess, organisation, variability of colour and density . 
\item[Tag genome:] a dataset with human ratings of the extent to which a number of tags apply to different movies and books. \citet{kumar-etal-2024-ranking} selected 38 tags for movies and 32 tags for books which can be viewed as ordinal features, all corresponding to adjectives (e.g.\ scary, quirky, suspenseful). The original movie ratings were obtained by \citet{DBLP:journals/tiis/VigSR12}, while the book ratings were obtained by \citet{DBLP:conf/chiir/KotkovMMSNG22}.
\item[Physical properties:] a dataset focused on three physical properties: mass, size and height. The data was originally created by \citet{DBLP:conf/corl/StandleySCS17} and \citet{liu-etal-2022-things}. It was used to evaluate LLMs by \citet{li-etal-2023-language-models} and subsequently cleaned by \citet{chatterjee-etal-2023-cabbage}, who removed 7 items.
\item[Wikidata:] a dataset with 20 numerical features obtained from Wikidata, collected by \citet{kumar-etal-2024-ranking} (e.g.\ the length of rivers, population of countries, and date of birth of people).
\end{description}

\noindent We will furthermore experiment on the following datasets, which have not yet been considered for evaluating LLMs, to the best of our knowledge:
\begin{description}
\item[Odour:] a dataset of 200 odorants collected by \citet{moss2016odorant}. A total of 103 participants rated odorants across nine dimensions. The authors reported that the following four were the most useful as normative data: familiarity, intensity, pleasantness, and irritability. We therefore also focus on these dimensions. 
\item[Music:] a dataset of 364 music excerpts from different genres,  collected by a panel of nine music experts \cite{strauss2024emotion}. The 517 participants rated the excerpts based on the emotions they felt, using the following dimensions from the Geneva Emotion Music Scale (GEMS) \cite{zentner2008emotions}: wonder, transcendence, tenderness, nostalgia, peacefulness, energy, joyful activation, sadness and tension.
\end{description}

\section{Experiments}
We refer to our proposed approach as \emph{ProtoSim} (Prototype Similarity).\footnote{Our code and preprocessed datasets are available at \url{https://github.com/niteshroyal/conceptual-spaces}.} ProtoSim is clearly more practical than prompting LLMs to provide pairwise judgments, especially when large numbers of entities need to be ranked. 
Our main research question is whether or not the increased convenience of ProtoSim comes with a trade-off on performance. 

\subsection{Experimental Setup}

\paragraph{Models} We experiment with LLMs of different sizes and from different families: Llama3-8B \cite{DBLP:journals/corr/abs-2407-21783}, Qwen3-8B and Qwen3-14B \cite{yang2025qwen3technicalreport}, Mistral-Nemo-12B, Mistral-Small-24B, OLMo2-7B, OLMo2-13B \cite{DBLP:journals/corr/abs-2501-00656} and Phi4-14B \cite{DBLP:journals/corr/abs-2412-08905}. We furthermore experiment with the following pre-trained embedding models: E5-Mistral-7B \cite{wang-etal-2024-improving-text}, LLM2Vec-Llama3-8B, LLM2Vec-Llama3-8B-Sup, and LLM2Vec-Mistral-7B \cite{DBLP:journals/corr/abs-2404-05961}. 
We evaluate all models in two settings. First, we fine-tune the LLMs and pre-trained embedding models using the strategy from Section \ref{secMethodology} (ProtoSim). Second, we fine-tune the LLMs as pairwise rankers, using the methodology from \citet{kumar-etal-2024-ranking}. 

\paragraph{Methodology}
We evaluate the following variants of the fine-tuning strategy from Section \ref{secMethodology}. \textbf{Pre-trained}: we use the model without any fine-tuning.
\textbf{Classification}: we only fine-tune the model with the classification dataset that was collected from GPT-4o (i.e.\ loss $\mathcal{L}_1$).
\textbf{Rank-perc}: we only fine-tune on the ranking datasets (i.e.\ loss $\mathcal{L}_2$). As fine-tuning data, we use all perceptual datasets (i.e.\ Taste, Rocks, Odour, Music), apart from the dataset that is being evaluated.
\textbf{Rank-full}: similar as before, but we fine-tune on all datasets (i.e.\ also on Tag Genome, Physical Properties and Wikidata), again excluding the dataset that is being evaluated.
\textbf{Class + rank-perc}: use both the \emph{Classification} and \emph{Rank-perc} losses.
\textbf{Class + rank-full}: use both the \emph{Classification} and \emph{Rank-full} losses.
For the pairwise approach, only the ranking datasets can be used, i.e.\ \emph{Rank-perc} and \emph{Rank-full}. However, we also report results for pre-trained models with pairwise few-shot prompting. The prompts we used for this purpose are included in Appendix \ref{sec: experimental details}.


\paragraph{Benchmarks}
The datasets discussed in Section \ref{secDatasets} are used for both training and testing, using a leave-one-out strategy. In particular, when testing on a given dataset, we train on all the other datasets in the case of \emph{rank-full} (and all the other perceptual datasets for \emph{rank-perc}). Our experiments thus focus on the ability of the models to generalize to different domains than the ones they have seen during training. The \emph{classification} dataset is open-domain, but this is a small dataset of 123 categories, which is not focused on perceptual properties and does not provide any information about ranking.


\subsection{Results}

\begin{table}[t]
\footnotesize
\centering
\setlength\tabcolsep{2.85pt}
\begin{tabular}{l ccccccc}
\toprule
& \rotatebox{90}{\textbf{Sweetness}} &  \rotatebox{90}{\textbf{Saltiness}} &  \rotatebox{90}{\textbf{Sourness}} &  \rotatebox{90}{\textbf{Bitterness}} &  \rotatebox{90}{\textbf{Umaminess}} &  \rotatebox{90} {\textbf{Fattiness}} &  \rotatebox{90} {\textbf{Average}}\\
\midrule
\multicolumn{8}{c}{\textsc{ProtoSim} (Llama3-8B)}\\
\midrule
Pre-trained &55.6	&	57.6	&	50.6	&	47.1	&	62.1	&	48.2	& 53.5\\ 
Classification & 77.6	&	78.8	&	\underline{\textbf{70.3}}	&	\underline{\textbf{64.4}}	&	70.3	&	72.6	& 72.4\\
Rank-perc & 77.9	&	75.3	&	56.5	&	55.9	&	68.5	&	63.2	& 66.2\\
Rank-full & 73.2	&	70.6	&	53.2	&	51.2	&	63.8	&	72.1	& 64.0\\
Class + rank-perc & \textbf{78.2}	&	\underline{\textbf{79.1}}	&	70.0	&	60.6	&	\underline{\textbf{72.9}}	&	75.0	& \underline{\textbf{72.6}}\\
Class + rank-full & 77.1	&	75.6	&	68.5	&	58.8	&	68.5	&	\underline{\textbf{76.2}}	& 70.8\\
\midrule
\multicolumn{8}{c}{\textsc{ProtoSim} (LLM2Vec-Llama3-8B-Sup)}\\
\midrule
Pre-trained & 70.0	&	57.1	&	62.7	&	48.5	&	57.7	&	60.9 & 59.5	\\
Classification & 76.2	&	74.1	&	\textbf{67.9}	&	\textbf{62.6}	&	\textbf{67.1}	&	70.0 & 69.7	\\
Rank-perc & 75.0	&	76.8	&	58.2	&	55.3	&	64.7	&	\textbf{70.9} & 66.8	\\
Rank-full & 72.6	&	72.9	&	55.9	&	51.8	&	58.2	&	70.3 & 63.6	\\
Class + rank-perc & \textbf{77.6}	&	\textbf{77.4}	&	66.8	&	61.2	&	66.2	&	70.3	& \textbf{69.9}\\
Class + rank-full & 76.2	&	76.2	&	65.0	&	61.2	&	66.5	&	69.4 & 69.1	\\
\midrule
\multicolumn{8}{c}{\textsc{Pairwise approach} (Llama3-8B)}\\
\midrule
Few-shot & 52.4	&	52.6	&	47.1	&	51.8	&	51.2	&	52.4 & 51.2	\\
Rank-perc & 55.3	&	62.9	&	56.8	&	55.3	&	52.1	&	57.4 & 56.6	\\ 
Rank-full & \underline{\textbf{79.7}}	&	\textbf{71.5}	&	\textbf{62.7}	&	\textbf{62.1}	&	\textbf{63.5}	&	\textbf{72.1}	& \textbf{68.6}\\
\bottomrule
\end{tabular}
\caption{Comparison of different fine-tuning strategies (accuracy \% on pairwise comparisons). The best results within each block are highlighted in bold. The best results overall are underlined.\label{tabComparisonTrainingStrategies}}
\end{table}

\paragraph{Comparing Fine-tuning Strategies}
We first determine the best fine-tuning strategy for each approach. For this analysis, we use Llama3-8B (for the variants based on pre-trained LLMs) and LLM2Vec-Llama3-8B-Sup (for the variants based on pre-trained embedding models). The results are summarized in Table \ref{tabComparisonTrainingStrategies} for the Taste dataset. For ProtoSim with Llama3-8B, we can clearly see the effectiveness of the classification dataset, enabling an increase from 53.5\% to 72.4\%. Despite its small size, it successfully allow us to align the embedding space of the entities with the embedding space of the prototypes. Only fine-tuning on the ranking objective also helps, but it underperforms the classification approach. The \emph{Class + rank-perc} approach overall performs best, outperforming \emph{Classification} in four of the six dimensions. For ProtoSim with LLM2Vec-Llama3-8B-Sup, the findings are broadly similar, with \emph{Class + rank-perc} again performing best. For the remainder of the experiments, we will therefore fix \emph{Class + rank-perc} as the fine-tuning strategy for the ProtoSim experiments. When it comes to the pairwise approach, \emph{Rank-full} outperforms \emph{Rank-perc}. In the following, we will thus fix \emph{Rank-full} as the fine-tuning strategy for the experiments with the pairwise approach.

\begin{table}[t]
\footnotesize
\centering
\setlength\tabcolsep{2.8pt}
\resizebox{\columnwidth}{!}{\begin{tabular}{l ccccccc}
\toprule
& \rotatebox{90}{\textbf{Sweetness}} &  \rotatebox{90}{\textbf{Saltiness}} &  \rotatebox{90}{\textbf{Sourness}} &  \rotatebox{90}{\textbf{Bitterness}} &  \rotatebox{90}{\textbf{Umaminess}} &  \rotatebox{90} {\textbf{Fattiness}} &  \rotatebox{90} {\textbf{Average}} \\
\midrule
\multicolumn{8}{c}{\textsc{ProtoSim (LLMs)}}\\
\midrule
Llama3-8B	&	\textbf{78.2} 	&	\underline{\textbf{79.1}} 	&	70.0 	&	60.6 	&	\underline{\textbf{72.9}}	&	75.0 &	\underline{\textbf{72.7}} 	\\
Qwen3-8B	&	75.9 	&	71.5 	&	63.2 	&	62.6 	&	61.5	&	72.7 	&	67.9 	\\
Qwen3-14B	&	74.7 	&	70.3 	&	66.2 	&	60.6 	&	63.4	&	72.4 	&	67.9 	\\
Mistral-12B	&	76.8 	&	72.9 	&	70.9 	&	64.1 	&	64.4	&	75.9 	&	70.8 	\\
Mistral-24B	&	77.9 	&	76.2 	&	70.3 	&	59.1 	&	62.7	&	74.7 	&	70.2 	\\
OLMo2-7B	&	75.0 	&	68.2 	&	\underline{\textbf{75.6}} 	&	\textbf{65.9} 	&	67.4	&	76.5 	&	71.4 	\\
OLMo2-13B	&	76.8 	&	70.0 	&	69.1 	&	63.8 	&	56.5	&	74.7 	&	68.5 	\\
Phi4-14B	&	75.9 	&	69.4 	&	67.4 	&	61.5 	&	65.0	&	\textbf{76.8} 	&	69.3 	\\
 
\midrule

\multicolumn{8}{c}{\textsc{ProtoSim (Fine-tuned embedding models)}}\\
\midrule
E5-Mistral-7B	&	74.7 	&	\textbf{77.1} 	&	64.4 	&	62.4 	&	62.9 	&	\textbf{75.6} 	&	69.5 	\\
LLM2Vec \scriptsize{(Llama3)}	&	\textbf{76.5} 	&	76.2 	&	\textbf{65.3} 	&	60.6 	&	66.8 	&	72.4 	&	\textbf{69.6} 	\\
LLM2Vec \scriptsize{(Mistral)}	&	71.5 	&	74.7 	&	62.4 	&	\textbf{65.0} 	&	\textbf{70.3} 	&	72.4 	&	69.4 	\\
\midrule
\multicolumn{8}{c}{\textsc{ProtoSim (Pre-trained embedding models)}}\\
\midrule
E5-Mistral-7B	&	\textbf{68.5} 	&	\textbf{63.5} 	&	\textbf{64.4} 	&	51.5 	&	\textbf{61.8} 	&	\textbf{65.0} 	&	\textbf{62.5} 	\\
LLM2Vec \scriptsize{(Llama3)}	&	\textbf{68.5} 	&	45.9 	&	52.4 	&	42.9 	&	55.0 	&	38.5 	&	50.5 	\\
LLM2Vec \scriptsize{(Mistral)}	&	65.3 	&	54.4 	&	58.8 	&	\textbf{64.4} 	&	51.2 	&	50.6 	&	57.5 	\\
\midrule
\multicolumn{8}{c}{\textsc{Pairwise approach}}\\
\midrule
Llama3-8B	&	\underline{\textbf{79.7}}&	71.5 	&	62.6 	&	62.1 	&	63.5 	&	72.1 	&	68.6 	\\
Qwen3-8B	&	78.5 	&	71.5 	&	63.8 	&	58.5 	&	65.0 	&	72.4 	&	68.3 	\\
Qwen3-14B	&	\underline{\textbf{79.7}} 	&	73.5 	&	61.5 	&	55.9 	&	64.7 	&	\textbf{77.6} 	&	68.8 	\\
Mistral-12B	&	79.4 	&	73.8 	&	\textbf{67.6} 	&	56.5 	&	63.5 	&	72.4 	&	68.9 	\\
Mistral-24B	&	76.8 	&	\textbf{77.4} 	&	66.2 	&	\underline{\textbf{67.6}} 	&	67.4 	&	75.9 	&	\textbf{71.9} 	\\
OLMo2-7B	&	74.1 	&	64.1 	&	60.0 	&	57.9 	&	62.4 	&	69.4 	&	64.7 	\\
OLMo2-13B	&	79.4 	&	71.8 	&	62.4 	&	64.4 	&	64.7 	&	70.6 	&	68.9 	\\
Phi4-14B	&	75.6 	&	68.2 	&	60.9 	&	57.1 	&	\textbf{70.6} 	&	69.1 	&	66.9 	\\
\midrule
\multicolumn{8}{c}{\textsc{Zero-shot LLMs}}\\
\midrule
GPT-4o	&	73.5 	&	73.2 	&	68.5 	&	56.8 	&	65.6 	&	74.4 	&	68.7 	\\
GPT-4.1	&	\textbf{79.4} 	&	\textbf{76.2} 	&	\textbf{71.2} 	&	\textbf{58.5} 	&	\textbf{70.3} 	&	\underline{\textbf{78.2}} 	&	\textbf{72.3} 	\\
\bottomrule
\end{tabular}}
\caption{Comparison of different models (accuracy \% on pairwise comparisons). The best results within each block are highlighted in bold. The best results overall are underlined. ProtoSim results are obtained with \texttt{Class + rank-perc}, results for the pairwise model are for \texttt{Class + rank-full}. \label{tabComparisonModels}}
\end{table}

\begin{table*}[t]
\footnotesize
\centering
\setlength\tabcolsep{2.5pt}

 \resizebox{\textwidth}{!}{\begin{tabular}{l ccccccc cccc cccccccccc}
\toprule
& \multicolumn{7}{c}{\textbf{Rocks}} & \multicolumn{4}{c}{\textbf{Odour}} & \multicolumn{9}{c}{\textbf{Music}}\\
\cmidrule(lr){2-8}\cmidrule(lr){9-12}\cmidrule(lr){13-21}
&  \rotatebox{90}{\textbf{Lightness}} &  \rotatebox{90}{\textbf{Grain size}} &  \rotatebox{90}{\textbf{Roughness}} &  \rotatebox{90}{\textbf{Shininess}} &  \rotatebox{90}{\textbf{Organisation}}  &  \rotatebox{90}{\textbf{Variability}} &  \rotatebox{90}{\textbf{Density}} &  \rotatebox{90}{\textbf{Familiarity}} & \rotatebox{90}{\textbf{Intensity}} & \rotatebox{90}{\textbf{Pleasantness}} & \rotatebox{90}{\textbf{Irritability}}
&  \rotatebox{90}{\textbf{Wonder}} &  \rotatebox{90}{\textbf{Transcendence}} &  \rotatebox{90}{\textbf{Tenderness}} &  \rotatebox{90}{\textbf{Nostalgia}} &  \rotatebox{90}{\textbf{Peacefulness}} &  \rotatebox{90}{\textbf{Energy}} &  \rotatebox{90}{\textbf{Joyful activation}} &  \rotatebox{90}{\textbf{Sadness}} &  \rotatebox{90}{\textbf{Tension}} &  \rotatebox{90}{\textbf{Average}}\\
\midrule
\multicolumn{22}{c}{\textsc{ProtoSim (LLMs)}}\\
\midrule
Llama3-8B &	79.7&	62.6&	60.3&	64.7&	59.4&	\textbf{68.2}&	78.8&		42.6&	\textbf{62.9}&	72.6&	64.4&		53.5&	61.2&	69.7&	60.0&	66.5&	60.0&	63.2&	60.0&	\textbf{64.7}&	63.8\\
Mistral-24B &	79.7&	70.6&	58.8&	64.1&	55.9&	60.3&	77.3&		\textbf{64.7}&	60.0&	\textbf{74.4}&	\textbf{66.2}&		53.8&	62.6&	69.1&	58.8&	65.9&	60.3&	60.9&	\textbf{65.0}&	63.2&	64.6\\
\midrule
\multicolumn{22}{c}{\textsc{Pairwise approach}}\\
\midrule
Llama3-8B &	79.4&	70.9&	60.9&	60.6&	58.2&	50.0&	69.7&		53.8&	52.1&	58.8&	56.8&		\textbf{59.1}&	57.9&	71.5&	59.4&	62.6&	59.7&	55.6&	64.7&	61.8&	61.2\\
Mistral-24B	& 79.7&	\textbf{80.0}&	62.6&	\textbf{67.9}&	50.3&	67.9&	78.0&		57.4&	53.8&	58.2&	59.1&		55.6&	60.9&	68.8&	52.1&	63.2&	58.8&	50.9&	62.1&	56.8&	62.2\\
\midrule
\multicolumn{22}{c}{\textsc{Zero-shot LLMs}}\\
\midrule
GPT-4o & 59.7 & 75.6 & 55.9 & 63.2 & \textbf{65.0} & 52.4 & 69.7 & 58.5 & 48.5 & 58.8 &	51.2 & 50.6 & \textbf{63.8} & 66.5 & 51.8 & 72.1 & 59.4 & 62.9 &	55.0 &	60.6 & 60.1\\
GPT-4.1 &	\textbf{80.6} &	77.6&	\textbf{68.5} &	67.1&	56.2&	61.5&	\textbf{84.8}&		58.5&	53.2&	72.1 &	62.4&		54.7&	61.8&	\textbf{75.3}&	\textbf{62.9}&	\textbf{75.0}&	\textbf{65.6}&	\textbf{63.8}&	60.6&	\textbf{64.7}&	\textbf{66.3}\\
\bottomrule
\end{tabular}}
\caption{Comparison of different models (accuracy \% on pairwise comparisons). The best overall results for each quality dimension are highlighted in bold. ProtoSim results are obtained with \texttt{Class + rank-perc}, results for the pairwise model are for \texttt{Class + rank-full}. \label{tabPerceptualDatasets}}
\end{table*}

\begin{table}[t]
\footnotesize
\centering
\setlength\tabcolsep{2.6pt}
\begin{tabular}{l cc cc ccc c}
\toprule
& \multicolumn{2}{c}{\textbf{WD}} & \multicolumn{2}{c}{\textbf{TG}} & \multicolumn{3}{c}{\textbf{Phys}}\\
\cmidrule(lr){2-3}\cmidrule(lr){4-5}\cmidrule(lr){6-8}
&  \rotatebox{90}{\textbf{WD1}} &  \rotatebox{90}{\textbf{WD2}} &  \rotatebox{90}{\textbf{Movies}} &  \rotatebox{90}{\textbf{Books}} &  \rotatebox{90}{\textbf{Size}}  &  \rotatebox{90}{\textbf{Mass}} &  \rotatebox{90}{\textbf{Height}} &  \rotatebox{90}{\textbf{Average}}\\
\midrule
\multicolumn{9}{c}{\textsc{ProtoSim (LLMs)}}\\
\midrule
Llama3-8B & 65.6&	68.6&	71.1&	61.6&	75.3&	58.4&	78.3&	68.4\\
Mistral-24B & 66.0&	71.6&	\textbf{72.5}&	58.0&	66.9&	53.6&	65.7&	64.9\\
\midrule
\multicolumn{9}{c}{\textsc{Pairwise approach}}\\
\midrule
Llama3-8B & 64.8&	58.6&	64.0&	51.6&	75.3&	59.6&	83.7&	65.4\\
Mistral-24B &65.4&	64.0&	62.6&	53.8&	88.0&	61.4&	92.2&	69.6\\
\midrule
\multicolumn{9}{c}{\textsc{Zero-shot LLMs}}\\
\midrule
GPT-4o & 68.0 & 79.2 &	67.4 &	61.1 &	92.2 &	50.0 &	85.5 &	71.9\\
GPT-4.1 & \textbf{81.0} &	\textbf{89.4} &	72.1&	\textbf{67.1}&	\textbf{98.2}&	\textbf{64.5}&	\textbf{97.0}&	\textbf{81.3}\\
\bottomrule
\end{tabular}
\caption{Comparison of different models (accuracy \% on pairwise comparisons). The best overall results for each quality dimension are highlighted in bold.  ProtoSim results are obtained with \texttt{Class + rank-perc}, results for the pairwise model are for \texttt{Class + rank-full}. 
\label{tabNonPerceptualDatasets}}
\end{table}

\begin{figure}[t]
\centering
\includegraphics[width=\columnwidth]{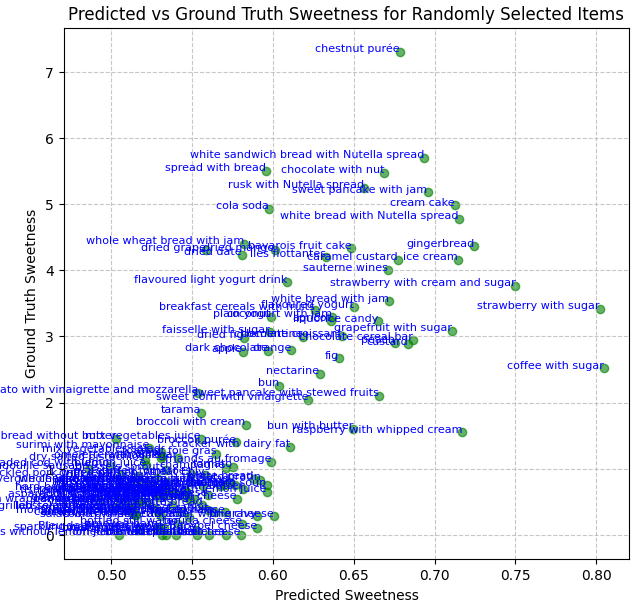}
\caption{Scatter plot showing the predicted sweetness of a food item (X-axis) and the ground truth rating (Y-axis).\label{figScatterPlotSweet}}
\end{figure}

\paragraph{Comparing Models}
Table \ref{tabComparisonModels} compares the performance of a number of different models, for each of the considered approaches. For this analysis, we still focus on the Taste dataset, and fix the fine-tuning strategies as explained above. For ProtoSim, Llama3-8B achieves the best results for three dimensions, with OLMo2-7B for two dimensions and Phi4-14B for one dimension. Surprisingly, increasing model size does not seem to improve results. For instance, the performance of Qwen3-8B and Qwen3-14B is almost identical, Mistral-12B outperforms Mistral-24B, and OLMo2-7B outperforms OLMo2-13B (on average). 
ProtoSim can be used with LLMs and with pre-trained embedding models. We might expect that starting from a model such as LLM2Vec would have some advantages, as the model has already been pre-trained to generate embeddings. However, we found such models to underperform Llama3-8B. When using the pre-trained embedding models without any fine-tuning, performance is substantially lower. In that case, we also see clear differences between E5 and the LLM2Vec models. However, after fine-tuning these differences disappear. 
When comparing ProtoSim and the pairwise approach, their relative performance depends on the LLM which is used. The best results overall are obtained by ProtoSim with Llama3-8B. ProtoSim is also better when Mistral-12B, OLMo2-7B or Phi4-14B is used. Conversely, the pairwise approach is better for the Qwen models, Mistral-24B and OLMo2-13B. 
Finally, we also report zero-shot results with GPT-4o and GPT-4.1 in the table. We found GPT-4.1 to consistently improve on GPT-4o, while performing slightly worse than ProtoSim with Llama3-8B on average.

\paragraph{Evaluation on Different Domains}
We now analyze the results on the other datasets. First, Table \ref{tabPerceptualDatasets} shows the results for the remaining perceptual datasets. As before, the ProtoSim models are trained using \textit{Class + rank-perc} and the pairwise models using \textit{Class + rank-full}, based on our findings from Table \ref{tabComparisonTrainingStrategies}. Based on the results from Table \ref{tabComparisonModels} we focus this analysis on Llama3-8B (as the best-performing model for ProtoSim and a representative smaller model) and Mistral-24B (as the best-performing model for the pairwise approach and a representative larger model). We find that ProtoSim outperforms the pairwise approach on average, although there is some variation between the three considered domains: the pairwise approach with Mistral-24B outperforms ProtoSim on Rocks; ProtoSim outperforms the pairwise approach on Odour, especially for Mistral-24B; and both approaches perform relatively similarly on Music, with ProtoSim being slightly better on average. We find that Mistral-24B outperforms Llama3-8B even for ProtoSim (especially for Odour), in contrast to our earlier findings on Taste. GPT-4.1 achieves the best results on Rocks and Music, but underperforms ProtoSim with Mistral-24B on Odour.

Table \ref{tabNonPerceptualDatasets} summarizes the results for the non-perceptual datasets. ProtoSim outperforms the pairwise approach on the Tag Genome dataset and, to a lesser extent, on Wikidata, but the pairwise approach performs better on Physical Properties. GPT-4.1 substantially outperforms the other methods on Wikidata and Physical properties, which are the two datasets that involve factual numerical attributes. For Tag Genome, which involves subjective labels, the performance of GPT-4.1 is more in line with ProtoSim and the pairwise approach. 

\subsection{Analysis}
\paragraph{Predicting Degrees of Sweetness}
For the main experiments, we have only focused on ranking. However, in contrast to the pairwise approach, ProtoSim associates a numerical score $\textit{emb}(\gamma(e))\cdot \textit{emb}(\gamma(f))$ with every entity $e$ and feature $f$, which we can interpret as the coordinate of a conceptual space dimension. As such, we can also use this method for predicting the \emph{degree} to which an entity has some feature. We analyze this for the particular example of \emph{sweetness} from the Food dataset. Figure \ref{figScatterPlotSweet} compares the predicted sweetness score with the ground truth sweetness values (which were obtained as the average sweetness rating that was assigned by all annotators). For this analysis, we have used the ProtoSim model with Llama3-8B (trained using \emph{class + rank-perc}). The figure shows a random sample of 150 food items. The figure shows a clear correlation between the predicted and ground truth scores (Pearson correlation for the full set of 590 food items: 0.752). In the bottom-left corner of the plot, we can see a large set of items which are considered to be clearly non-sweet, both by the human annotators and by the model. The items that are rated to be sweetest by the human annotators are all predicted to be sweet by the model as well (with \emph{chestnut pur\'ee} as an outlier).
However, food items with intermediate levels of sweetness can be more challenging. For instance, \emph{coffee with sugar} is far less sweet than predicted by the model, while \emph{cola soda} and \emph{whole wheat bread with jam} are sweeter than predicted. 

A similar analysis for the other perceptual dimensions can be found in Appendix \ref{appendixPredictingRatings}.

\paragraph{Qualitative Analysis}
To better understand which kinds of features can be modelled using ProtoSim, we carried out a qualitative analysis using a question-answering dataset about recipes \cite{zhang2023recipe}. Each question specifies a preference for a particular type of food (e.g.\ \emph{a quick breakfast for a rushed school morning}), and the task is to select the most appropriate option among 5 listed alternatives. We select the option whose embedding is most similar to the stated preference. We found that the model was generally able to handle a variety of commonsense properties (e.g.\ \emph{a toddler-friendly fried snack for a birthday party}). However, we also noticed three key limitations: difficulties with negative preferences, being overly sensitive to lexical overlap, and sometimes focusing too much on one aspect of the query. A detailed analysis can be found in Appendix \ref{appendixExtendedErrorAnalysis}.

\section{Conclusions}
We have shown that LLM embeddings can serve as conceptual space representations of perceptual features. While previous work had already shown the potential of LLMs for modelling perceptual features, this was based on pairwise comparison prompts, which are not practical when representations for large numbers of entities are needed. To model a given quality dimension (e.g.\ sweetness) we obtain an LLM embedding of a corresponding prototype description (e.g.\ ``a sweet food''). The main idea is that we can then simply compare this embedding with the embeddings of the entities of interest. However, we found this to perform poorly with pre-trained LLMs (including LLM-based embedding models), due to the fact that the embeddings of the prototype descriptions and the entities are not aligned. To address this, we align the embeddings by fine-tuning the LLM on a small synthetically generated dataset. After this alignment step, we found the proposed strategy to be highly effective, matching and often even surpassing the performance of the pairwise approach. 

\section*{Acknowledgments}
This work was supported by EPSRC grant EP/W003309/1.

\section*{Limitations}
The problem of aligning vector spaces has been extensively studied within the context of cross-lingual word embeddings \cite{DBLP:journals/corr/MikolovLS13,xing-etal-2015-normalized,artetxe-etal-2020-cross}. Such methods essentially learn a linear transformation to align two monolingual vector spaces. It is possible that a similar approach might be affective for aligning prototype and entity embedding spaces we well, which would mean that the fine-tuning step could be avoided. Apart from being more efficient (e.g.\ in terms of storing model parameters), this might also help to prevent any catastrophic forgetting. However, in preliminary experiments (presented in Appendix \ref{appendixAdditionalExperiments}) we failed to obtain competitive results with this approach. A further investigation into the potential of linear mappings is left for future work.


In our experiments, we have focused on ranking, rather than measuring the degree to which features are satisfied. We illustrated the potential of our model to predict degrees of sweetness, and a similar analysis for the other quality dimensions can be found in Appendix \ref{appendixPredictingRatings}, but a formal evaluation is left for future work.  More generally, conceptual spaces are commonly used for evaluating similarity. For instance, we expect that a learned conceptual space of taste, composed of the six considered taste dimensions, would allow us to estimate human similarity judgments more reliably than is possible with the original LLM embeddings. Note that the problem of estimating similarity judgments can also be related to the problem of estimating causal inner products in LLM embeddings spaces \cite{DBLP:conf/icml/ParkCV24}.

\bibliography{anthology,custom}

\appendix

\begin{table*}
\footnotesize
\centering
\begin{tabular}{lll}
\toprule
\textbf{Model Name} & \textbf{Hugging Face URL} & \textbf{License}\\
\midrule
Llama3-8B& \url{meta-llama/Meta-Llama-3-8B} & Llama 3\\
Qwen3-8B&  \url{Qwen/Qwen3-8B} & Apache 2.0\\
Qwen3-14B &\url{Qwen/Qwen3-14B} & Apache 2.0\\
Mistral-Nemo-12B & \url{mistralai/Mistral-Nemo-Base-2407} & Apache 2.0\\
Mistral-Small-24B & \url{mistralai/Mistral-Small-24B-Base-2501} & Apache 2.0\\
OLMo2-7B & \url{allenai/OLMo-2-1124-7B} & Apache 2.0\\
OLMo2-13B & \url{allenai/OLMo-2-1124-13B} & Apache 2.0\\
Phi4-14B & \url{microsoft/phi-4} & MIT \\
\midrule
E5-Mistral-7B & \url{intfloat/e5-mistral-7b-instruct} & MIT\\
LLM2Vec-Llama3-8B & \url{McGill-NLP/LLM2Vec-Meta-Llama-3-8B-Instruct-mntp} & MIT\\
LLM2Vec-Llama3-8B-Sup & \url{McGill-NLP/LLM2Vec-Meta-Llama-3-8B-Instruct-mntp-supervised} & MIT\\
LLM2Vec-Mistral-7B & \url{McGill-NLP/LLM2Vec-Mistral-7B-Instruct-v2-mntp} & MIT\\
\bottomrule
\end{tabular}
\caption{Details of the models used in the experiments. \label{tabModelsURLs}}
\end{table*}

\section{Experimental Details} \label{sec: experimental details}
\paragraph{Models} Table \ref{tabModelsURLs} provides the details of the models that were used in our experiments. Experiments with GPT-4o and GPT-4.1 were carried out using the OpenAI API\footnote{\url{https://platform.openai.com}}. We used versions \texttt{gpt-4o-2024-11-20} and \texttt{gpt-4.1-2025-04-14} respectively.

\paragraph{Fine-tuning Methodology} To fine-tune the base models, we used the QLoRa method, which allows converting the floating-point 32 format to smaller data types. In particular, for all the models, we used 4-bit quantization for efficient training. In the QLoRa configuration, $r$ (the rank of the low-rank matrix used in the adapters) was set to 32,  $\alpha$ (the scaling factor for the learned weights) was set to 64, and dropout was set to 0.05. We applied QLoRa to all the linear layers of the models, including \textit{q\_proj}, \textit{k\_proj}, \textit{v\_proj}, \textit{o\_proj}, \textit{gate\_proj}, \textit{up\_proj}, \textit{down\_proj}, and \textit{lm\_head}. In all our experiments, the temperature parameter $T$ was set to $0.25$ for the classification loss, the scaling factor $\alpha$ was set to $10$, and $\lambda$ was set to $0.25$.

\paragraph{Computing Infrastructure} The fine-tuning experiments were conducted on a workstation equipped with an NVIDIA GeForce RTX 4090 GPU with 24GB of VRAM.

\paragraph{Prompts} For the few-shot configuration in Table \ref{tabComparisonTrainingStrategies}, we used the following prompt with three in-context demonstrations:

\begin{lstlisting}
The task is to answer questions that involve comparing perceptual features of two entities. Please answer with Yes or No only. In the worst case, if you do not know the answer then choose randomly between Yes and No.

This question is about two surfaces: Is mirror more reflective than still water surface?
Yes
This question is about two materials: Is silk fabric more lustrous than polished metal?
No
This question is about two sounds: Is operatic aria more melodious than car alarm?
Yes
\end{lstlisting}

\smallskip
\noindent We used the following prompt for the experiments with GPT-4o and GPT-4.1:

\begin{lstlisting}
Answer the following with Yes or No only. In the worst case, if you don't know the answer then choose randomly between Yes and No.
\end{lstlisting}



\begin{table*}
\footnotesize
\centering
\begin{tabular}{lp{160pt}p{127pt}}
\toprule
\textbf{Target Property} & \textbf{Examples} & \textbf{Negative Properties}\\   
\midrule
long river & Nile River, Amazon River, Yangtze River, Yenisei River, Yellow River, Ob-Irtysh River, Congo River & short river, polluted river, dry river, small city\\
\midrule
influential artist &  Pablo Picasso, Leonardo da Vinci, Vincent van Gogh, Claude Monet, Michelangelo, Rembrandt, Andy Warhol & unknown artist, amateur artist, unpopular artist, dry river\\
\midrule
loyal dog & German Shepherd, Labrador Retriever, Golden Retriever, Collie, Boxer, Beagle, Bulldog & independent dog, aloof dog, aggressive dog, small city\\
\midrule
energy efficient appliance & LED Light Bulbs, Smart Thermostats, Energy Star Refrigerators, Dual Flush Toilets, Solar Panels, High-Efficiency Washers, Electric Vehicles & high-energy consumption appliance, inefficient lighting, old model refrigerators, mild spice\\
\midrule
water sport & Swimming, Water Polo, Diving, Synchronized Swimming, Rowing, Canoeing, Surfing & land sport, winter sport, individual sport, dry desert\\
\midrule
transparent material & Glass, Acrylic, Polycarbonate, Quartz Crystal, Diamond, Clear Resin, Sapphire Crystal & opaque material, metallic material, porous material, poisonous flower\\
\midrule
rail transportation & Train, Tram, Monorail, Subway, High-speed Rail, Funicular, Light Rail & air transport, road transport, water transport, ancient language\\
\midrule
international law & Geneva Conventions, United Nations Charter, Hague Convention, UNCLOS, Treaty of Rome, Kyoto Protocol, Vienna Convention & domestic law, criminal law, civil law, ballroom dance\\
\midrule
domesticated animal & Dog, Cat, Horse, Cow, Sheep, Goat, Chicken & wild animal, exotic animal, marine animal, modern software architecture\\
\midrule
metaphysics philosophical branch &  Ontology, Cosmology, Theology, Epistemology, Phenomenology, Existentialism, Dualism & logic, ethics, aesthetics, binary mathematical operation\\
\midrule
acidic chemical compound & Hydrochloric Acid, Sulfuric Acid, Acetic Acid, Citric Acid, Nitric Acid, Phosphoric Acid, Carbonic Acid & basic compound, neutral compound, alkaline compound, military alliance\\
\midrule
phonological linguistic phenomenon & Assimilation, Elision, Lenition, Vowel Harmony, Consonant Mutation, Metathesis, Assimilation & syntactic phenomenon, semantic phenomenon, morphological feature, freshwater ecosystem\\
\bottomrule
\end{tabular}
\caption{Examples from the fine-tuning dataset that was collected using GPT-4o. \label{tabExamplesClassificationDataset}}
\end{table*}

\section{Fine-tuning Dataset} \label{appFinetuningDataset}


The fine-tuning dataset for classification was synthetically generated using GPT-4o. We provided a few manually created examples and asked GPT-4o to generate additional similar datapoints. Each datapoint was manually checked, and GPT-4o was also prompted to re-examine the datapoints it generated as part of the quality assurance process. Multiple prompts were used interactively to guide the model in generating datapoints that cover diverse domains. In total, 517 datapoints were generated; however, we randomly selected 123 datapoints to be used for fine-tuning, as the model was overfitting to this dataset when the full set of 517 data points were used. Table \ref{tabExamplesClassificationDataset} shows some examples of data points from the dataset.

\section{Evaluation Datasets}
For Taste, Rocks, Tag Genome, Physical Properties and Wikidata, we use the preprocessed datasets from \citet{kumar-etal-2024-ranking}, which are available from \url{https://github.com/niteshroyal/RankingUsingLLMs}. For the Odour and Music datasets, we obtained the datasets from the original publications. In particular, the Odour dataset is available as supplemental data at \url{https://www.frontiersin.org/journals/psychology/articles/10.3389/fpsyg.2016.01267/full}. The Music dataset is available from \url{https://osf.io/7ptmd/}.

For the Taste, Rocks and Physical properties datasets, we could not find any information about licensing. The Tag Genome dataset was released under CC BY-NC 3.0. Wikidata is available under a CC0 license. The Odour dataset was released under a CC BY 4.0 license.


\begin{figure}
\centering
\fbox{
\begin{minipage}[t]{0.94\linewidth}
\footnotesize
\noindent \textbf{Query}: \emph{a quick breakfast for a rushed school morning.}\\
\noindent \textbf{Options}:\\[-2em]
\begin{enumerate}
\setlength{\itemsep}{-3pt}
\setlength{\itemindent}{-3pt}
\item  \green{\textbf{Any cereal with milk}}
\item Eggs benedict - poached eggs, prosciutto on top of English muffins topped with a creamy Hollandaise sauce
\item Instant ramen with eggs, spinach and pickled cabbage
\item Breakfast pizza with sausage, cheddar, sour cream and jalapenos
\item Classic salted french fries made of only potatoes
\end{enumerate}
\end{minipage}}
\caption{Example question from the recipe dataset.\label{figRecipeExample}}
\end{figure}

\section{Qualitative Analysis}\label{appendixExtendedErrorAnalysis}
The dataset from \citet{zhang2023recipe} contains 500 multiple-choice questions, each with 5 alternatives. To evaluate our models, we first converted each question to a descriptive phrase (expressing the same preference as the original question) using GPT-4o. Figure \ref{figRecipeExample} shows a problem instance from the resulting dataset.

We first evaluated a number of LLMs on the original question answering benchmark, using a zero-shot prompt, achieving 91.4\% accuracy with GPT-4o and 89.4\% with Llama3-8B. This shows that, while many of the instances appear challenging, LLMs are generally capable of identifying the correct option.  We then tested our Llama3-8B ProtoSim model (fine-tuned without the taste dataset), as follows. We used the descriptive version of the query as the verbalization of the property. The five options are treated as the verbalization of entities. We then simply predict the option whose embedding is closest to the embedding of the query. The accuracy of this approach was 67.6\%.

Analyzing the results, we noticed that the model generally performs well on commonsense properties. For instance, the following queries were all answered correctly: (i) a quick breakfast for a rushed school morning, (ii) a toddler-friendly fried snack for a birthday party, (iii) diabetes-friendly cookies. However, Tables \ref{tabDetailedAnalysis}, \ref{tabAnalysisLexOveralp} and \ref{tabAnalysisNegation} illustrate three types of common errors that are made by the model (ProtoSim with Llama3-8B).








Table \ref{tabDetailedAnalysis} shows examples where the model focuses too much on one particular aspect of the specification. In the first example, the words \emph{post-cardio} and \emph{muscle} lead the model to select the \emph{protein smoothie} option, despite the fact that the description was asking for a \emph{snack}. Similarly, in the second example, the word \emph{antioxidants} leads to the model to the vitamin-rich smoothie, ignoring the fact that the query was asking for a \emph{salad}.


In Table \ref{tabAnalysisLexOveralp}, it is evident that the model is distracted by the lexical overlap between the query and some of the options. In the first example, the model selects an option that mentions \emph{brown rice}, which also occurs in the query, despite the fact that the chosen option is not a dessert. Similarly, due to significant lexical overlap with the final option, the model fails to acknowledge the term \emph{green} in the second example. 
In the final example, the model chose \emph{Low fat crab chowder made with imitation crabmeat and different vegetables} over the correct option \emph{Lighter clam chowder with bacon and vegetables, made with milk instead of cream} due to the presence of the words \emph{low fat} and \emph{chowder}, which also occur in the query. 

Table \ref{tabAnalysisNegation} illustrates how the model struggles to handle negative requirements, such as \textit{without cranberry sauce}, \textit{non-greasy} or \textit{lactose-free}. Such negative requirements can be critically important for recommendation systems \cite{wang2023learning}, but they are challenging to capture with embeddings.

\section{Predicting Numerical Ratings}\label{appendixPredictingRatings}
\begin{figure}[t]
\centering
\includegraphics[width=\columnwidth]{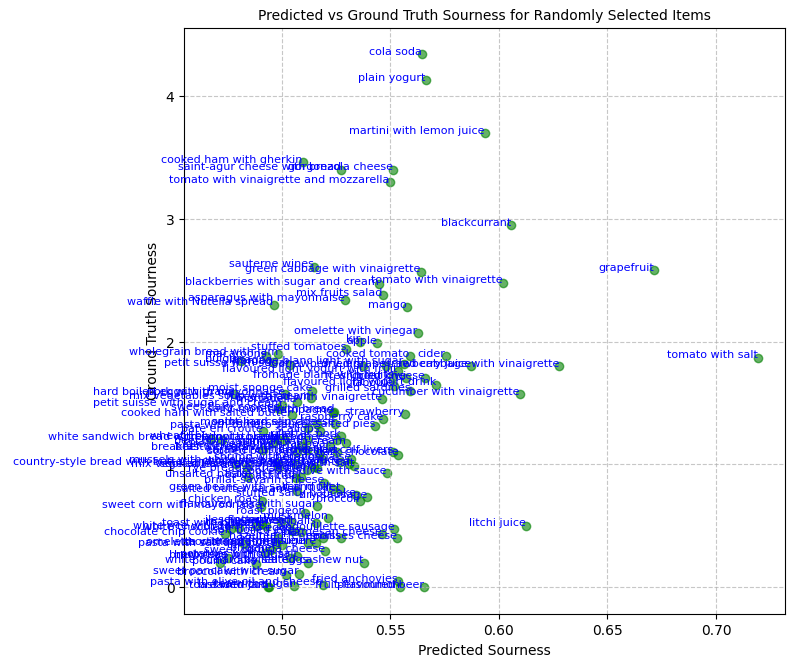}
\caption{Scatter plot showing the predicted sourness of a food item (X-axis) and the ground truth rating (Y-axis).\label{figScatterPlotTasteSour}}
\end{figure}

\begin{figure}[t]
\centering
\includegraphics[width=\columnwidth]{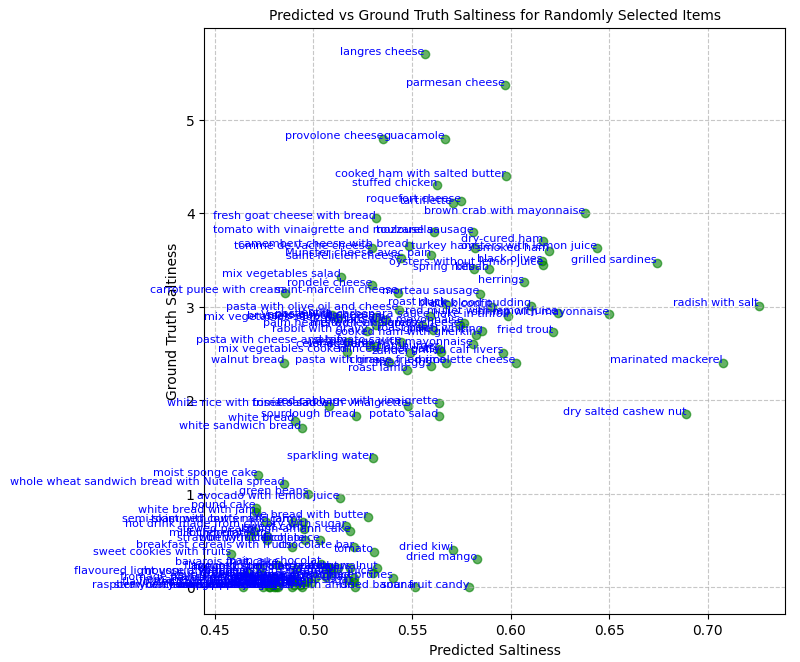}
\caption{Scatter plot showing the predicted saltiness of a food item (X-axis) and the ground truth rating (Y-axis).\label{figScatterPlotTasteSalty}}
\end{figure}

\begin{figure}[t]
\centering
\includegraphics[width=\columnwidth]{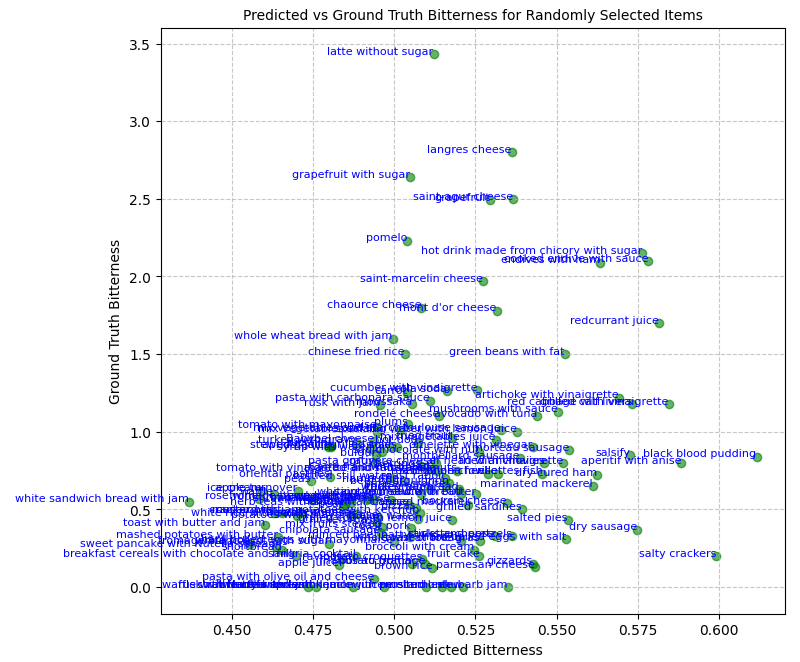}
\caption{Scatter plot showing the predicted bitterness of a food item (X-axis) and the ground truth rating (Y-axis).\label{figScatterPlotTasteBitter}}
\end{figure}

\begin{figure}[t]
\centering
\includegraphics[width=\columnwidth]{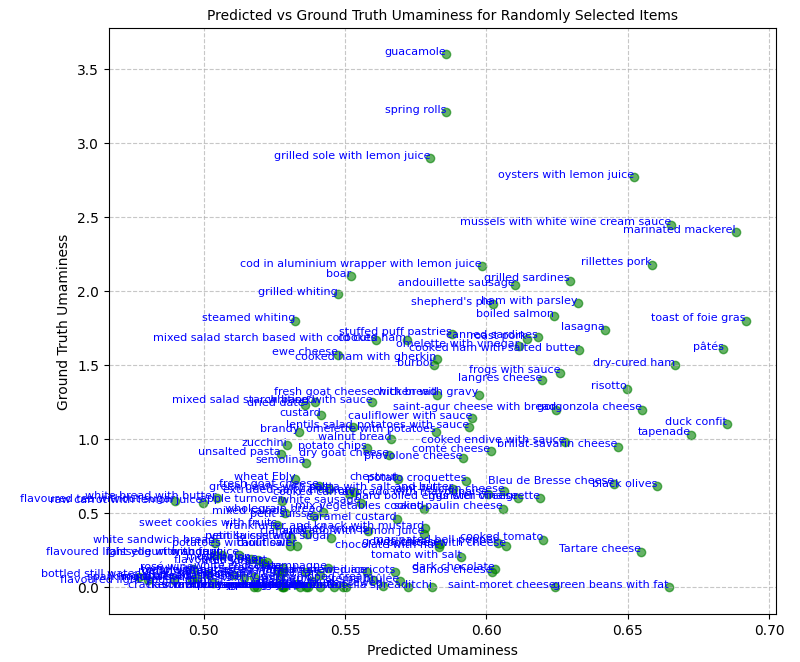}
\caption{Scatter plot showing the predicted umaminess of a food item (X-axis) and the ground truth rating (Y-axis).\label{figScatterPlotTasteUmami}}
\end{figure}

\begin{figure}[t]
\centering
\includegraphics[width=\columnwidth]{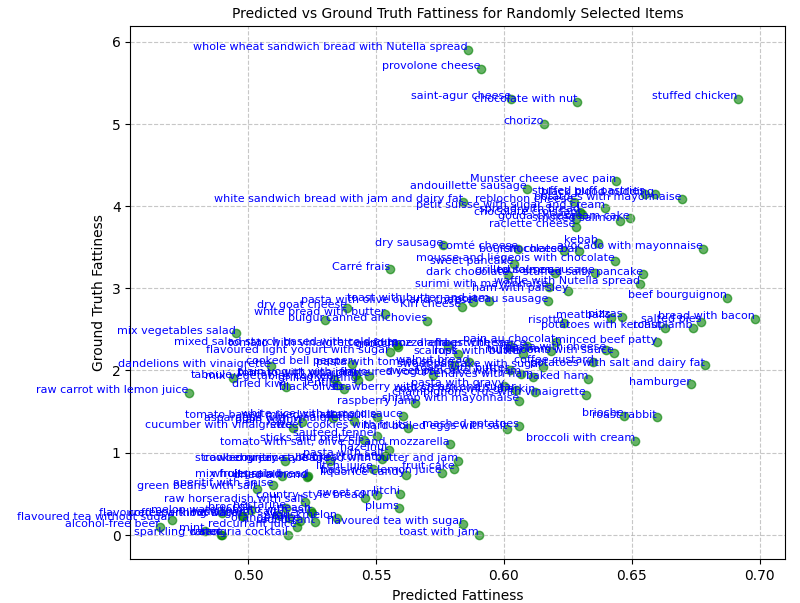}
\caption{Scatter plot showing the predicted fattiness of a food item (X-axis) and the ground truth rating (Y-axis).\label{figScatterPlotTasteFatty}}
\end{figure}

\begin{figure}[t]
\centering
\includegraphics[width=\columnwidth]{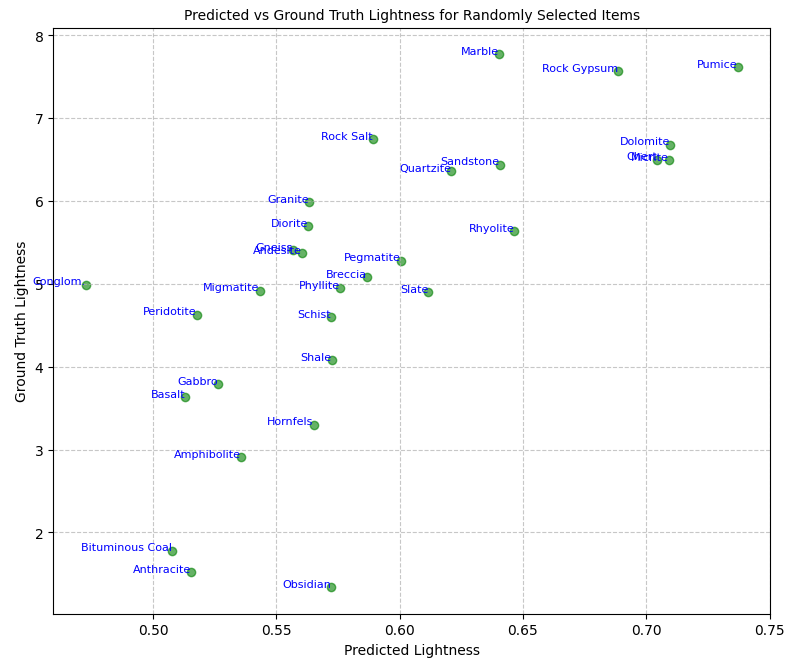}
\caption{Scatter plot showing the predicted lightness of rocks (X-axis) and the ground truth rating (Y-axis).\label{figScatterPlotRockLightness}}
\end{figure}

\begin{figure}[t]
\centering
\includegraphics[width=\columnwidth]{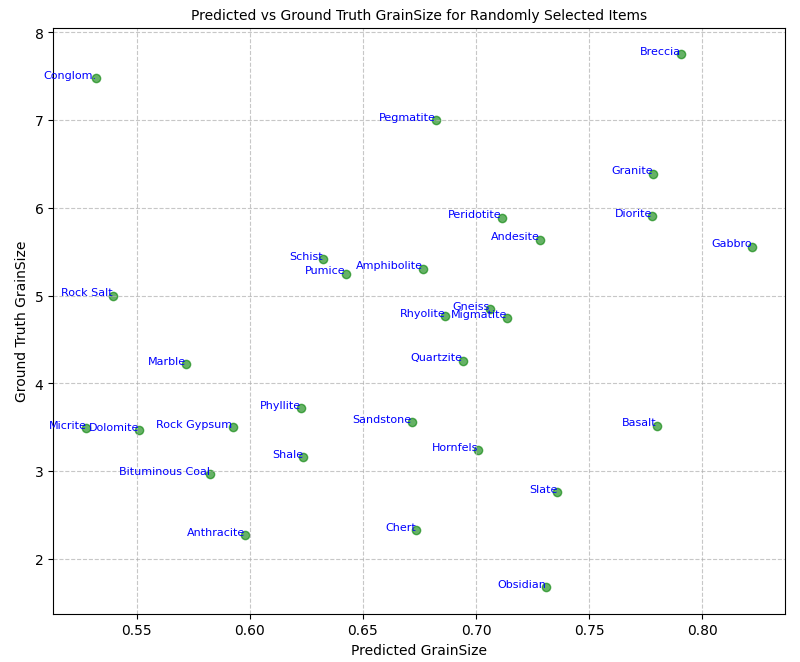}
\caption{Scatter plot showing the predicted grain size of rocks (X-axis) and the ground truth rating (Y-axis).\label{figScatterPlotGrainSize}}
\end{figure}

\begin{figure}[t]
\centering
\includegraphics[width=\columnwidth]{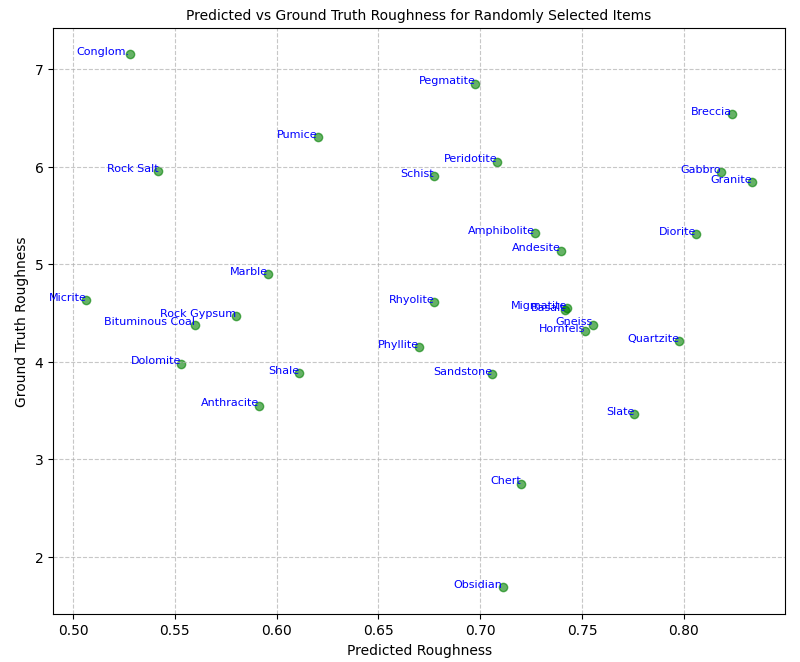}
\caption{Scatter plot showing the predicted roughness of rocks (X-axis) and the ground truth rating (Y-axis).\label{figScatterPlotRoughness}}
\end{figure}

\begin{figure}[t]
\centering
\includegraphics[width=\columnwidth]{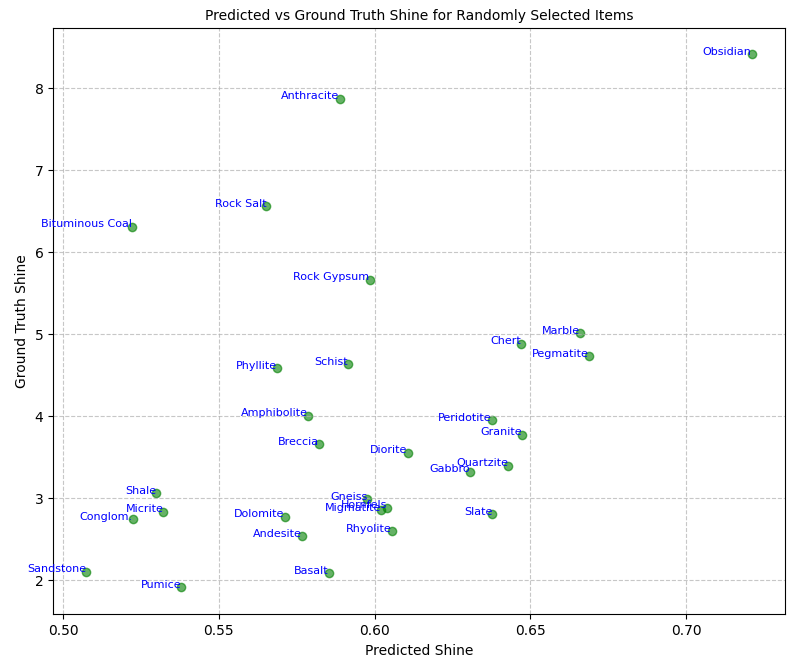}
\caption{Scatter plot showing the predicted shine of rocks (X-axis) and the ground truth rating (Y-axis).\label{figScatterPlotShine}}
\end{figure}

\begin{figure}[t]
\centering
\includegraphics[width=\columnwidth]{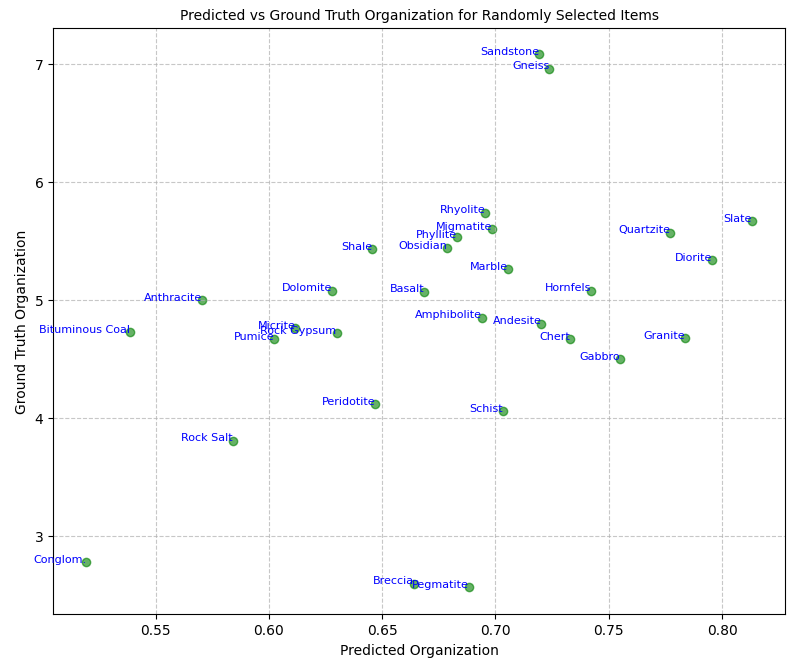}
\caption{Scatter plot showing the predicted organization of rocks (X-axis) and the ground truth rating (Y-axis).\label{figScatterPlotOrganization}}
\end{figure}

\begin{figure}[t]
\centering
\includegraphics[width=\columnwidth]{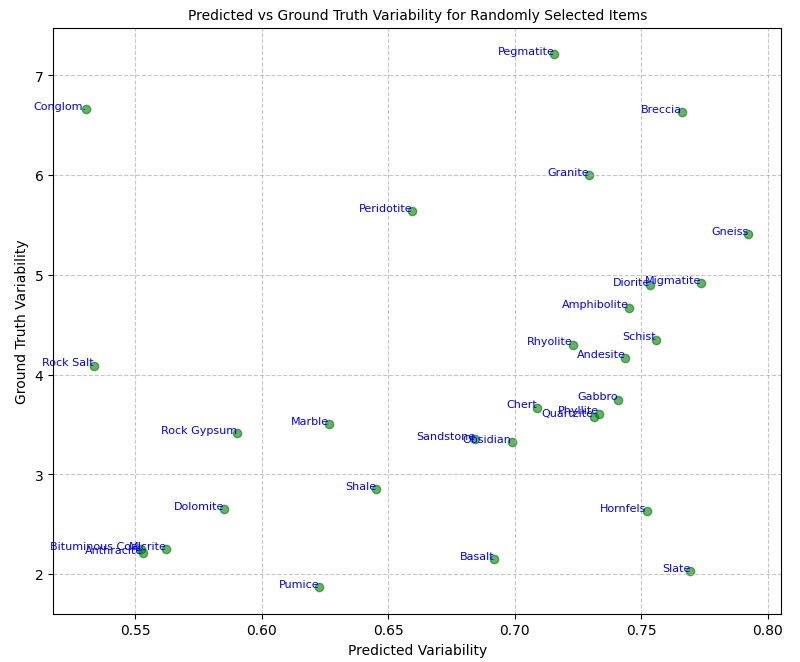}
\caption{Scatter plot showing the predicted variability of rocks (X-axis) and the ground truth rating (Y-axis).\label{figScatterPlotVariability}}
\end{figure}

\begin{figure}[t]
\centering
\includegraphics[width=\columnwidth]{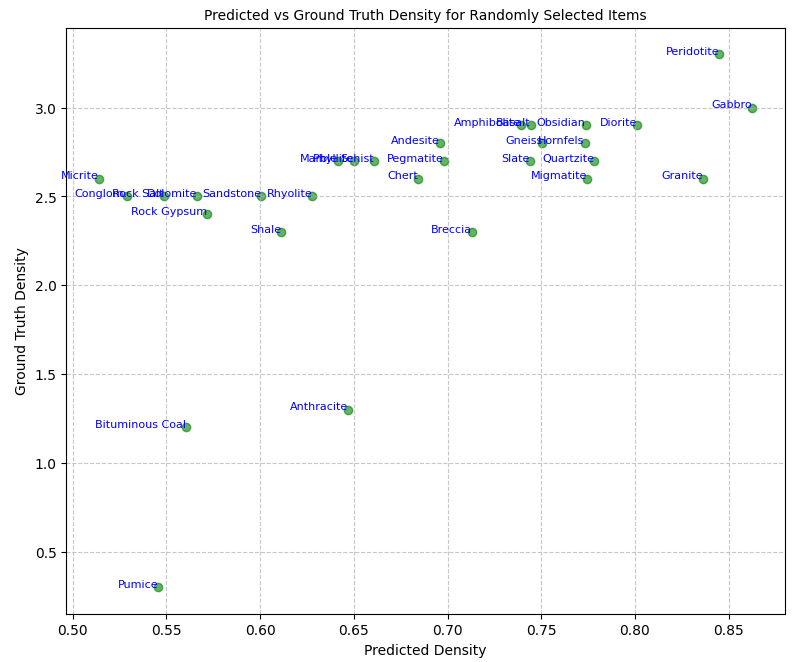}
\caption{Scatter plot showing the predicted density of rocks (X-axis) and the ground truth rating (Y-axis).\label{figScatterPlotDensity}}
\end{figure}

\begin{figure}[t]
\centering
\includegraphics[width=\columnwidth]{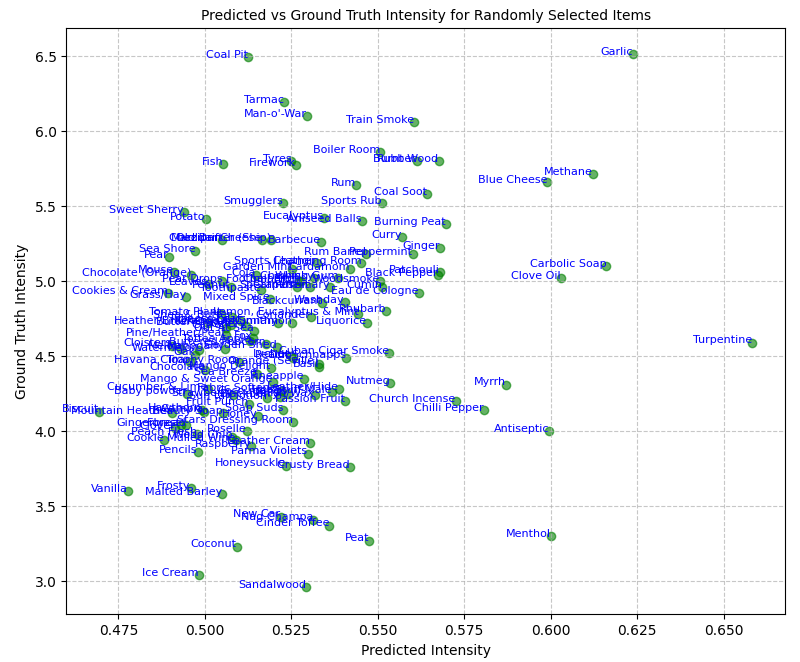}
\caption{Scatter plot showing the predicted intensity of odours (X-axis) and the ground truth rating (Y-axis).\label{figScatterPlotIntensity}}
\end{figure}

\begin{figure}[t]
\centering
\includegraphics[width=\columnwidth]{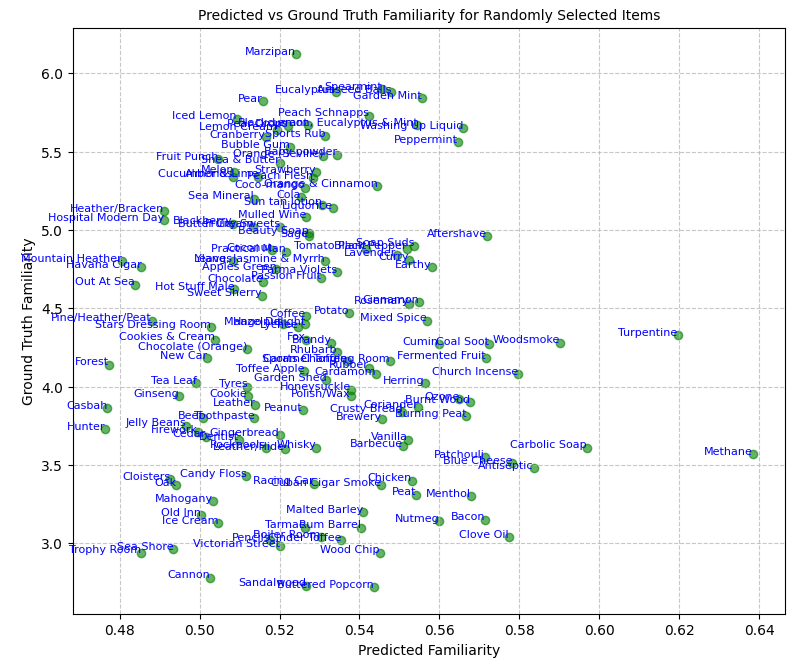}
\caption{Scatter plot showing the predicted familiarity of odours (X-axis) and the ground truth rating (Y-axis).\label{figScatterPlotFamiliar}}
\end{figure}

\begin{figure}[t]
\centering
\includegraphics[width=\columnwidth]{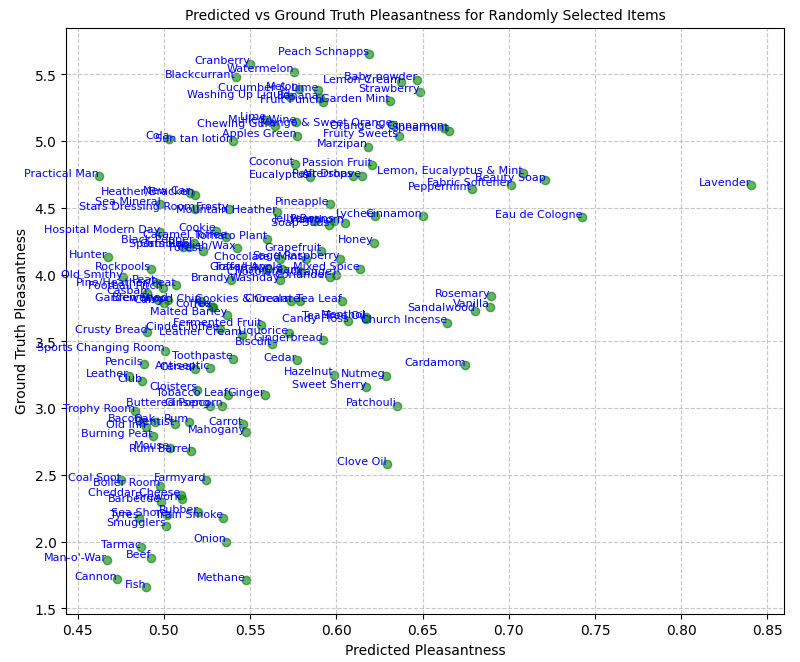}
\caption{Scatter plot showing the predicted pleasantness of odours (X-axis) and the ground truth rating (Y-axis).\label{figScatterPlotPleasant}}
\end{figure}

\begin{figure}[t]
\centering
\includegraphics[width=\columnwidth]{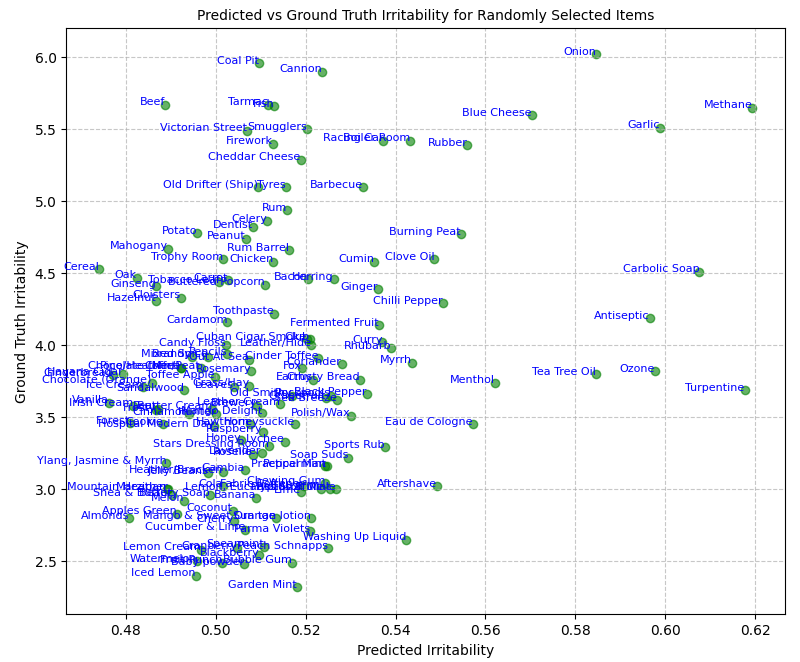}
\caption{Scatter plot showing the predicted irritability of odours (X-axis) and the ground truth rating (Y-axis).\label{figScatterPlotIrritability}}
\end{figure}

\begin{figure}[t]
\centering
\includegraphics[width=\columnwidth]{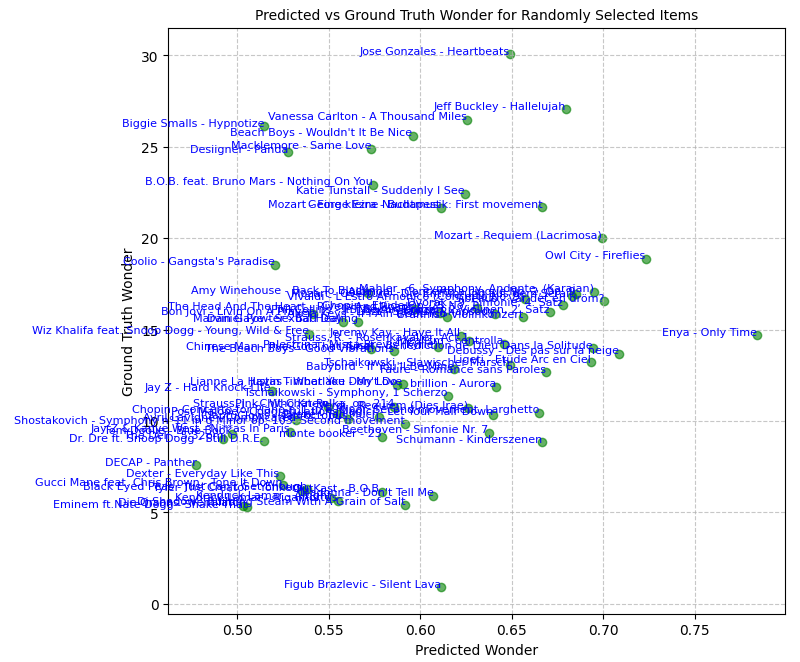}
\caption{Scatter plot showing the predicted feelings of wonder that music evokes (X-axis) and the ground truth rating (Y-axis).\label{figScatterPlotWonder}}
\end{figure}

\begin{figure}[t]
\centering
\includegraphics[width=\columnwidth]{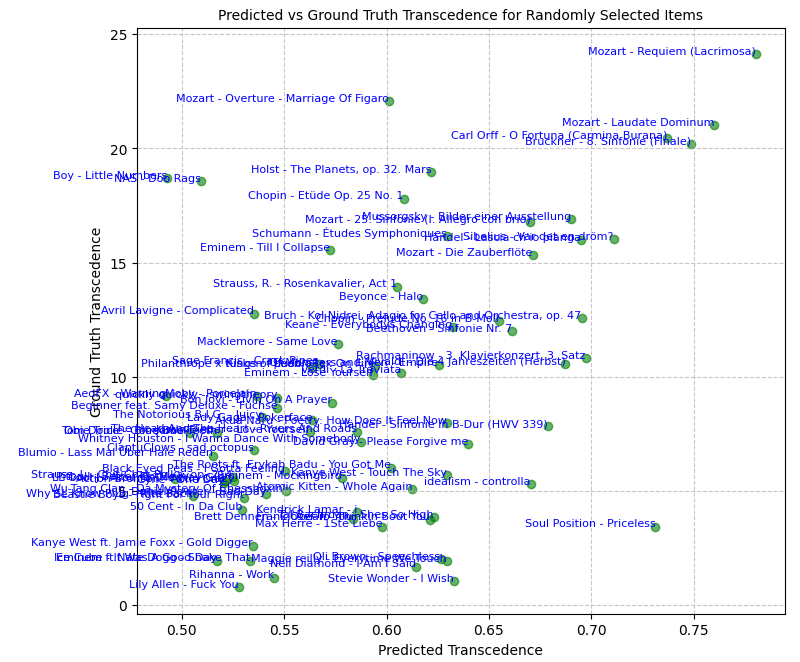}
\caption{Scatter plot showing the predicted feelings of transcendence that music evokes (X-axis) and the ground truth rating (Y-axis).\label{figScatterPlotTrans}}
\end{figure}

\begin{figure}[t]
\centering
\includegraphics[width=\columnwidth]{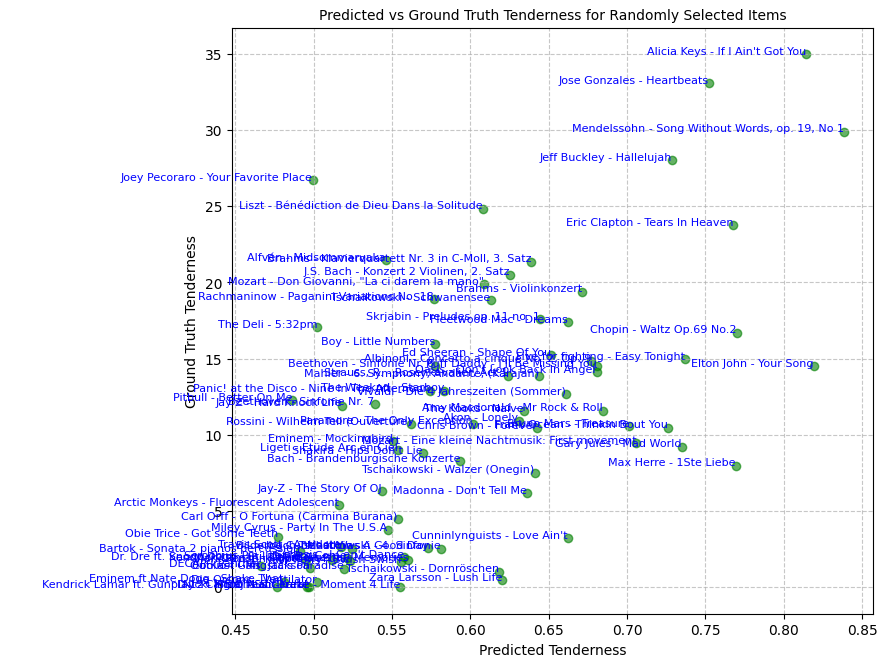}
\caption{Scatter plot showing the predicted feelings of tenderness that music evokes (X-axis) and the ground truth rating (Y-axis).\label{figScatterPlotTender}}
\end{figure}

\begin{figure}[t]
\centering
\includegraphics[width=\columnwidth]{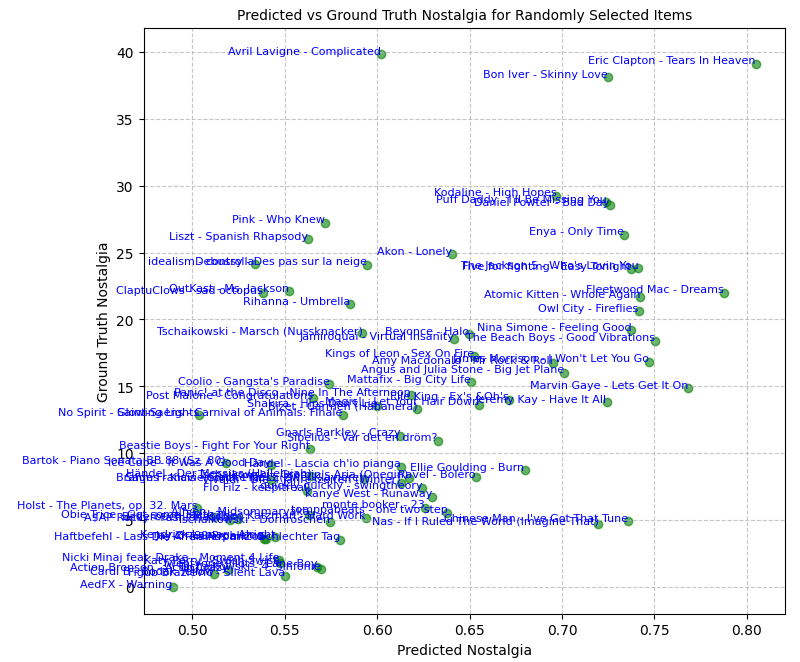}
\caption{Scatter plot showing the predicted feelings of nostalgia that music evokes (X-axis) and the ground truth rating (Y-axis).\label{figScatterPlotNostalgia}}
\end{figure}

\begin{figure}[t]
\centering
\includegraphics[width=\columnwidth]{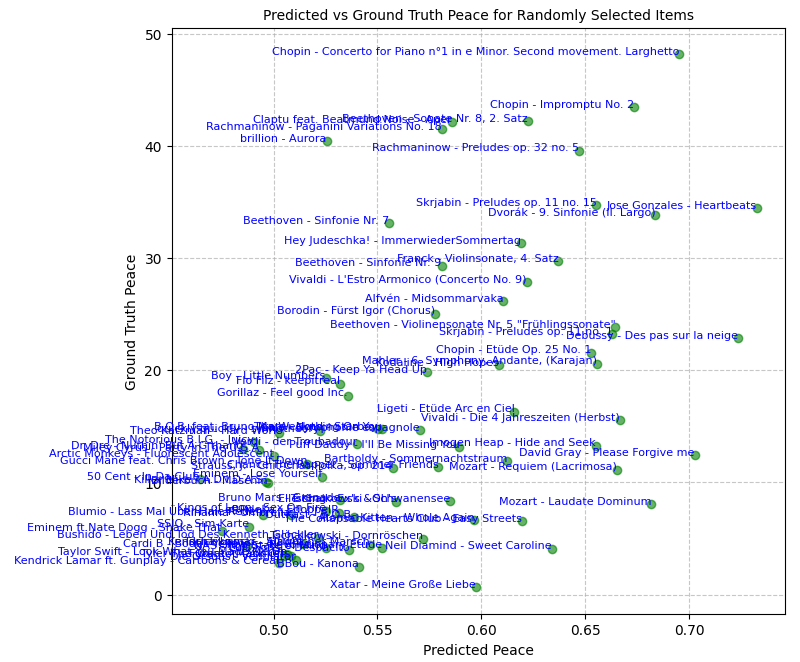}
\caption{Scatter plot showing the predicted feelings of peace that music evokes (X-axis) and the ground truth rating (Y-axis).\label{figScatterPlotPeace}}
\end{figure}

\begin{figure}[t]
\centering
\includegraphics[width=\columnwidth]{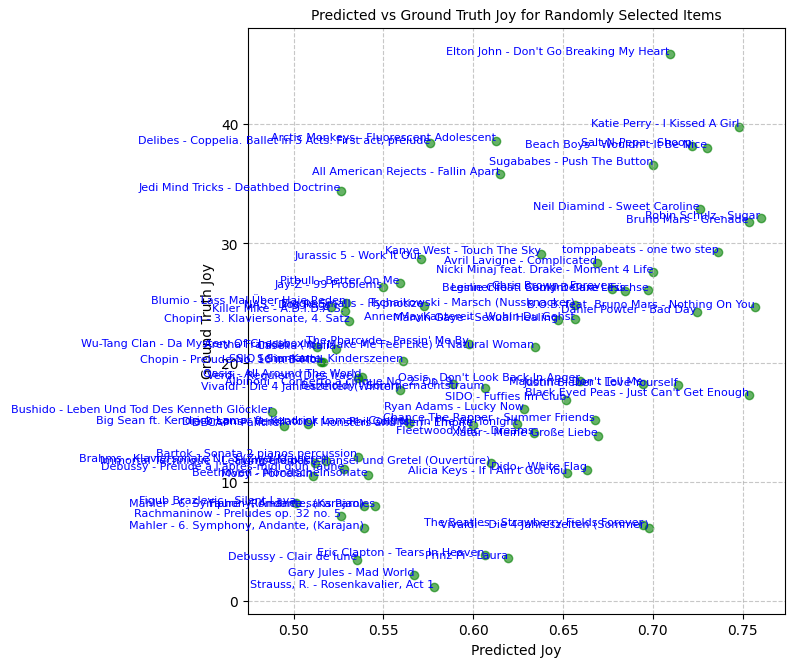}
\caption{Scatter plot showing the predicted feelings of joy that music evokes (X-axis) and the ground truth rating (Y-axis).\label{figScatterPlotJoy}}
\end{figure}

\begin{figure}[t]
\centering
\includegraphics[width=\columnwidth]{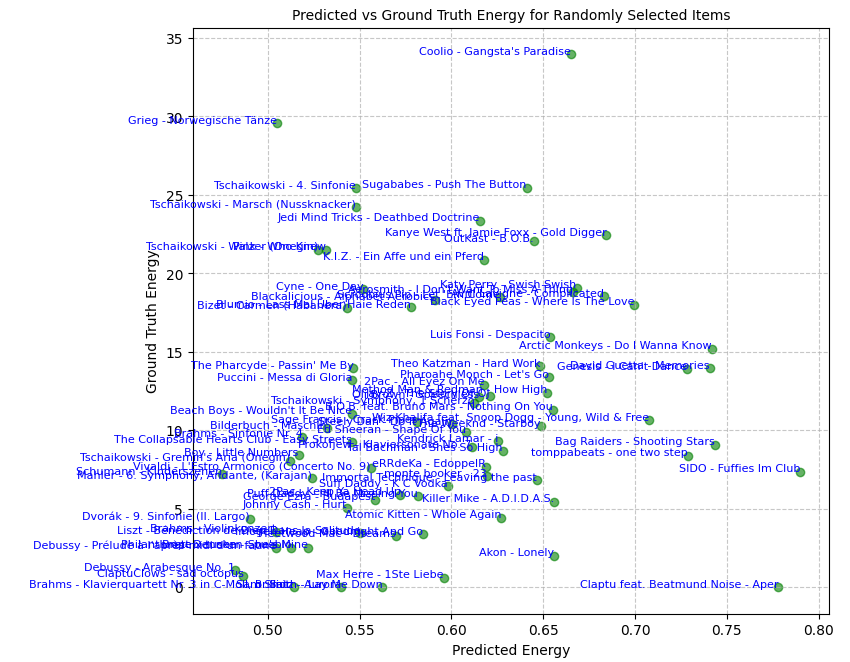}
\caption{Scatter plot showing the predicted feelings of energy that music evokes (X-axis) and the ground truth rating (Y-axis).\label{figScatterPlotEnergy}}
\end{figure}

\begin{figure}[t]
\centering
\includegraphics[width=\columnwidth]{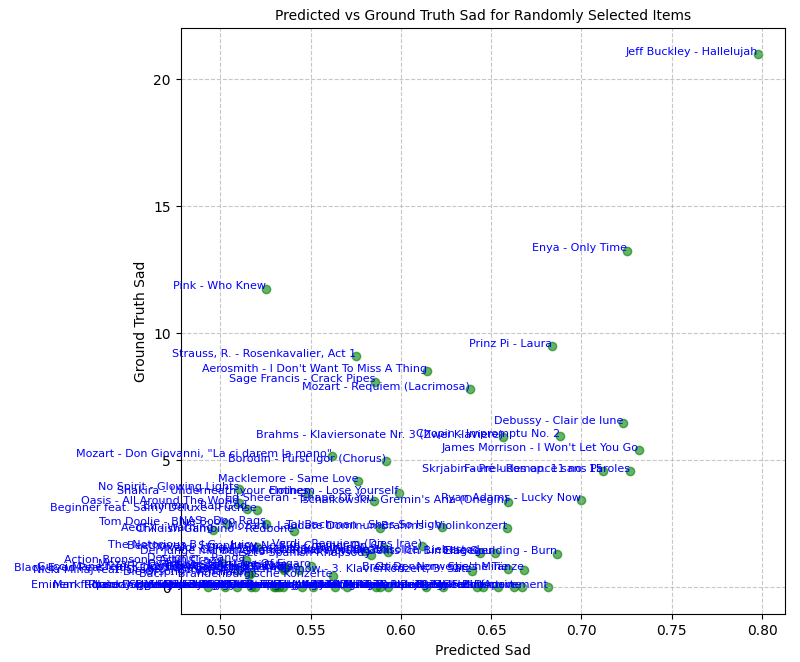}
\caption{Scatter plot showing the predicted feelings of sadness that music evokes (X-axis) and the ground truth rating (Y-axis).\label{figScatterPlotSad}}
\end{figure}

\begin{figure}[t]
\centering
\includegraphics[width=\columnwidth]{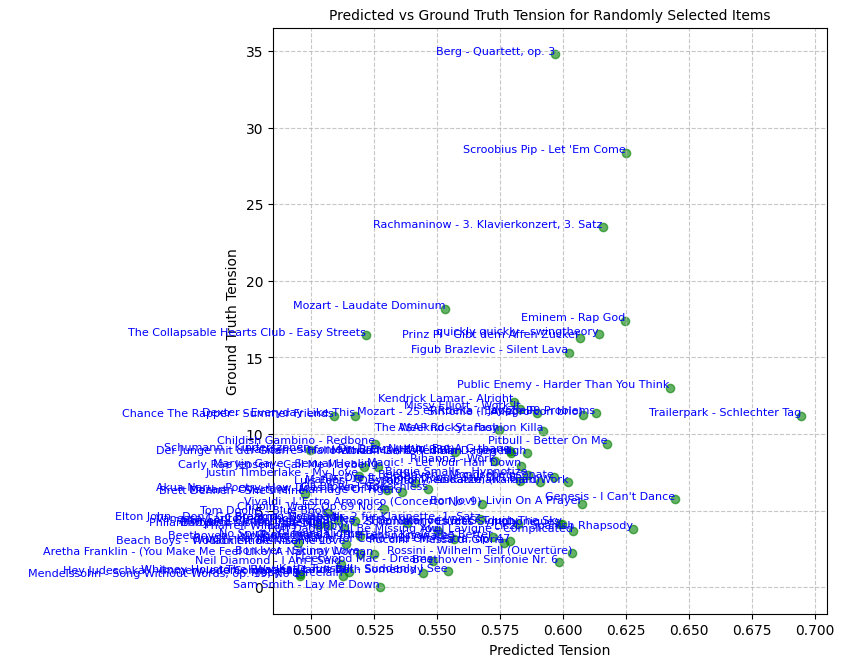}
\caption{Scatter plot showing the predicted feelings of tension that music evokes (X-axis) and the ground truth rating (Y-axis).\label{figScatterPlotTension}}
\end{figure}

In the main paper, we presented an analysis of the linear correlation between the predicted scores and the ground truth human ratings, for the sweetness dimension. Here we extend this analysis to the remaining dimensions from the taste domain, as well as the other perceptual domains.
Figures \ref{figScatterPlotTasteSour}--\ref{figScatterPlotTasteFatty} compare the predicted scores with the ground truth ratings for the remaining taste dimensions: sourness, saltiness, bitterness, fattiness and umaminess. Each figure shows a random sample of 150 food items. Similarly, 
Figures \ref{figScatterPlotRockLightness}--\ref{figScatterPlotDensity} show the analysis for Rocks, Figures \ref{figScatterPlotIntensity}--\ref{figScatterPlotIrritability} show the analysis for Odour (for a sample of 100 food and non-food items), and Figures \ref{figScatterPlotWonder}--\ref{figScatterPlotTension} show the results for Music (for a sample of 80 music titles).

Consistent with our findings in the main paper (and the insights from the qualitative analysis), we can see several cases where the predicted scores depend too much on a single word, or sub-phrase. For instance, the predicted saltiness of \emph{radish with salt} and \emph{dry salted cashew nuts} is too high (Figure \ref{figScatterPlotTasteSalty}), and similar for the predicted fattiness of \emph{broccoli with cream} (Figure \ref{figScatterPlotTasteFatty}). In the music domain, we see that \emph{soul position -- priceless} is incorrectly predicted to have a high value for \emph{transcendence} (Figure \ref{figScatterPlotTrans}), presumably due to the semantic relatedness of the words \emph{soul} and \emph{transcendence}.

Table \ref{tabPearsonCorrTaste} shows the Pearson correlations between the predicted scores and the ground truth ratings for the 6 dimensions from the food domain (computed w.r.t.\ the complete set of 590 food items). The Pearson correlations for the other domains are shown in Table \ref{tabPearsonCorrelationDatasets}. In all domains, we can see significant variation across dimensions. For Taste, sweetness and saltiness show reasonably strong correlation, while the correlation for bitterness is much lower. For Rocks, lightness is captured well by the model, but for most dimensions the correlation is weak. The Odour dataset is also challenging, with weak correlations on all dimensions apart from pleasantness. Finally, for music, we see moderate correlations for most of the dimensions.


\begin{table*}[t]
\centering
\footnotesize
\caption{Error analysis of the ProtoSim model. The table shows examples where the model focuses too much on one particular aspect of the query. The incorrect option chosen by the model is highlighted in red.\label{tabDetailedAnalysis}}
\begin{tabular}{p{120pt}p{240pt}}
\toprule
\textbf{Recipe Query} & \textbf{ProtoSim response}\\
\midrule
post-cardio snacks 
&1.Fruit salad with peaches, blackberries, strawberries and lime\\
for lean muscle maintenance &\red{2.Strawberry and banana protein smoothie}\\
&3.Classic chicken tenders - deep fried boneless chicken strips\\
&4.Fragrant pilaf made from quinoa\\
&5.Stir fried Japanese Shirataki noodles (low calorie noodles)\\
\midrule
{a salad rich with antioxidants}. 
& 1.Potato salad with extra virgin olive oil dressing\\ &2.Vitamin-rich soup made with vegetables\\
&\red{3.Vitamin-rich smoothies made with cranberries, carrot, mango, strawberries, and cantaloupe}\\
&4.Easy chicken legs made with Italian salad dressing\\
&5.Caesar salad dressing recipe made from scratch using raw cashews\\
\bottomrule
\end{tabular}
\end{table*}

\begin{table*}[t]
\centering
\footnotesize
\caption{Error analysis of the ProtoSim model. The table shows examples where the model relies too much on lexical overlap. The incorrect option chosen by the model is highlighted in red. \label{tabAnalysisLexOveralp}}
\begin{tabular}{p{120pt}p{240pt}}
\toprule
\textbf{Recipe Query} & \textbf{ProtoSim response}\\
\midrule
a dessert made with brown rice
& 1. Blueberry crisp containing blueberries, brown rice, rice bran, and walnuts\\
& 2. Long-grain white rice dish with onions\\
&3.Jasmine rice cooked with coconut milk\\
&\red{4.Brown rice and mushrooms cooked with vegetable stock, olive oil, and rice vinegar}\\
&5.Dessert treat made with butter, mini marshmallows, and Rice Krispie cereal\\
\midrule
{a post-workout green smoothie}
& 1.Garden veggie smoothie containing tomatoes, celery, parsley, and spinach\\ 
&2.Green chili made with bell peppers, beef stew meat, and chili peppers\\
&3.Pineapple smoothie containing buttermilk\\
&4.Frittata containing onions, zucchini, squash, red peppers, broccoli, and cauliflower\\
&\red{5.Berry post workout smoothie containing fresh raspberries strawberries, blueberries, and bananas}\\
\midrule
{a low-fat clam chowder recipe}
& 1.Lighter clam chowder with bacon and vegetables, made with milk instead of cream\\ 
&\red{2.Low fat crab chowder made with imitation crabmeat and different vegetables}\\
&3.Creamy linguine noodles with clams and onions\\
&4.Clam chowder made with half-and-half cream\\
&5.Clam chowder made with heavy whipping cream\\
\bottomrule
\end{tabular}
\end{table*}

\begin{table*}[t]
\centering
\footnotesize
\caption{Error analysis of the ProtoSim model. The table shows examples where the model fails to interpret negative requirements. The incorrect option chosen by the model is highlighted in red.\label{tabAnalysisNegation}}
\begin{tabular}{p{120pt}p{240pt}}
\toprule
\textbf{Recipe Query} & \textbf{ProtoSim response}\\
\midrule
grandma's thanksgiving dinner 
 & 1. Roast turkey with plum sauce\\
without cranberry sauce &\red{2. Roast turkey with sweet cranberry sauce}\\
&3.Baked chicken drumsticks in tomato sauce\\
&4.Chinese style crispy roast duck with hoisin sauce\\
&5.Classic seasoned roast beef with red pepper flakes\\
\midrule
{solid, non-greasy food} 
 & 1.Toast with seasonings\\ 
for a severe hangover &2.Pizza margherita - basic pizza with tomato sauce and mozzarella cheese\\
&3.Hot dogs with hot pepper sauce and green chillies\\
&4.Chickpea and mexican chilli soup\\
&\red{5.Miso based Shijimi clam broth for hangover prevention}\\
\midrule
{a quick, lactose-free}
 & 1.Boiled oats made with water\\ 
breakfast recipe &2.Oats boiled in milk\\
&\red{3.Microwaved oatmeal in milk}\\
&4.Milk boiled oats with cheese and syrup\\
&5.Enchiladas containing breakfast sausage, cheddar cheese, and a variety of vegetables\\
\bottomrule
\end{tabular}
\end{table*}

\section{Additional Experiments}\label{appendixAdditionalExperiments}

\paragraph{Linear Mapping}
In the main paper, we fine-tuned the LLM encoder to align the representations of entities and prototypes. Here we consider the alternative of keeping the LLM encoder frozen and instead learning a mapping from the entity and prototype embeddings onto a shared space.
In particular, following the cross-lingual embedding alignment literature \cite{artetxe2018robust}, we estimated an orthogonal linear mapping between prototype embeddings and centroid embeddings 
(using the standard Procrustes-based approach), and applied it to map prototypes onto the centroid space. Concretely, given prototype vectors $P \in \mathbb{R}^{n \times d}$ and  centroid vectors $C \in \mathbb{R}^{n \times d}$, we solve
\[
W^{*} = \arg\min_{W \in O(d)} \, \lVert P W - C \rVert_{F},
\]
where $O(d)$ is the set of orthogonal matrices. The optimal $W$ is obtained via singular value decomposition of $C^{\top} P$. At evaluation, prototype embeddings are multiplied by $W^{*}$. We evaluated this baseline on the Taste dataset under the same configuration as in Table \ref{tabComparisonTrainingStrategies}. Results are reported in Table \ref{tabComparisonLinearMapping}, where we used 5,670 prototypes generated by GPT-4.1 to learn the orthogonal mapping. The results are consistently lower than those of our main approach.

\begin{table}[t]
\footnotesize
\centering
\setlength\tabcolsep{2.85pt}
\begin{tabular}{l ccccccc}
\toprule
& \rotatebox{90}{\textbf{Sweetness}} &  \rotatebox{90}{\textbf{Saltiness}} &  \rotatebox{90}{\textbf{Sourness}} &  \rotatebox{90}{\textbf{Bitterness}} &  \rotatebox{90}{\textbf{Umaminess}} &  \rotatebox{90} {\textbf{Fattiness}} &  \rotatebox{90} {\textbf{Average}}\\
\midrule
Linear Mapping &59.7	&	55.9	&	57.7	&	47.9	&	58.5	&	47.9	& 54.6\\ 
\bottomrule
\end{tabular}
\caption{Results for the linear mapping approach on the Taste dataset. \label{tabComparisonLinearMapping}}
\end{table}

\paragraph{Significance Testing Results}
Our McNemar significance tests on the Taste dataset show that the ranking+classification objective does not yield consistent improvements over classification-only (Table \ref{tab:mcnemar}). For \textit{bitterness}, it significantly underperforms ($p<0.01$), while for \textit{fatiness} and \textit{umaminess} it achieves higher accuracy but without statistical significance (\textit{umaminess} is borderline, $p \approx 0.057$). For the remaining subsets, the differences are negligible. It thus remains unclear whether the additional ranking objective can bring meaningful improvements, compared to only using the classification objective.

\begin{figure*}[t]
\centering
\includegraphics[width=1.75\columnwidth]{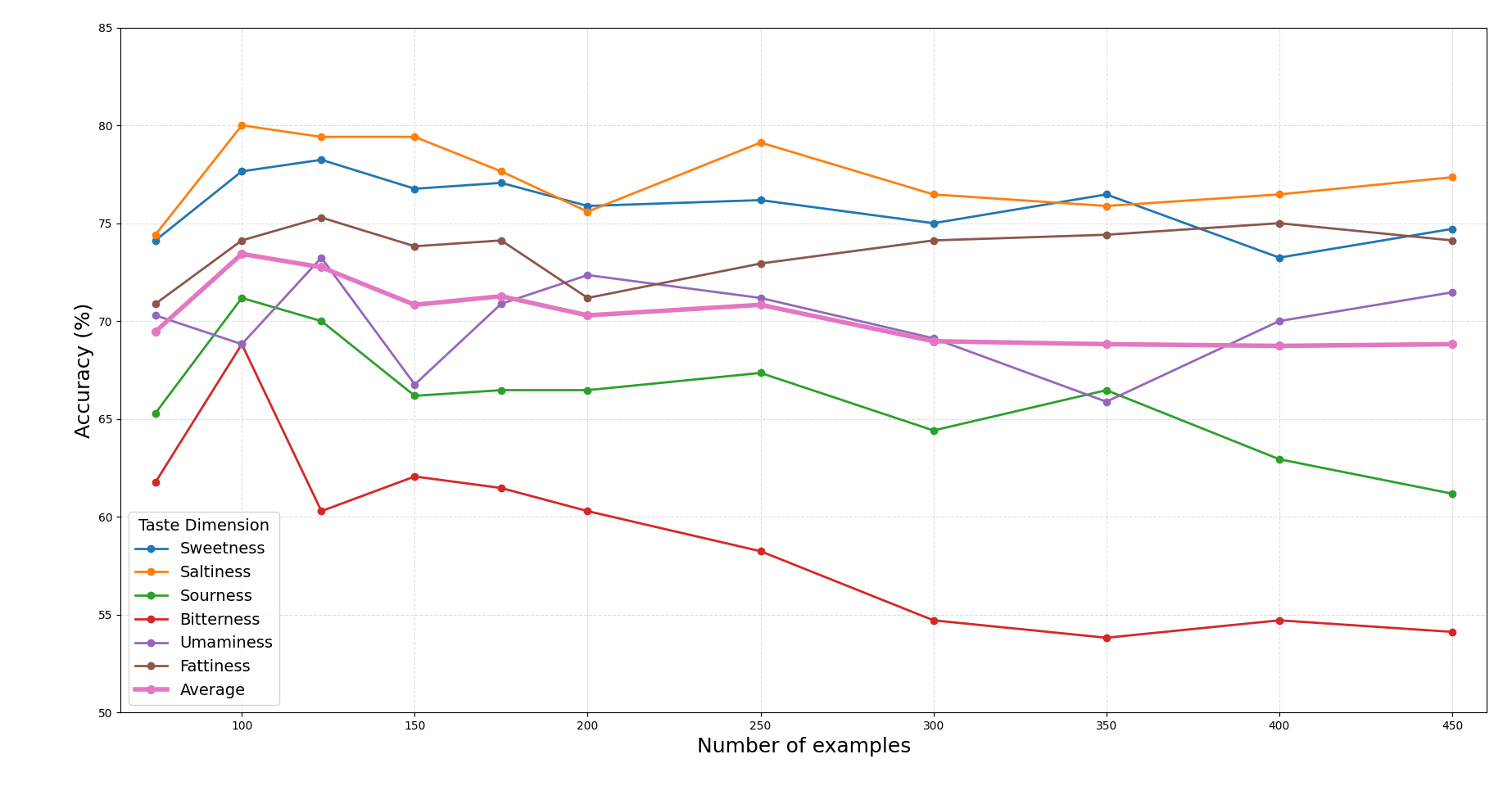}
\caption{Accuracy of \textsc{ProtoSim} (Llama3-8B) using the \emph{Class + rank-perc} approach on the Taste dataset, in function of the number of training examples used for classification.\label{figNumOfTrainingDatapoints}}
\end{figure*}

\paragraph{Effect of Classification Training Size}
We now analyze the effect of changing the number of training examples for the classification objective. For this experiment, we used the best-performing approach (\emph{Class + rank-perc}) and focused on the Taste dataset. Figure \ref{figNumOfTrainingDatapoints} plots the accuracy that was obtained, in function of the number of training datapoints. Interestingly, the results show that the accuracy peaks for relatively small training sizes (around 100–150 datapoints) and decreases when more training examples are used. 
A possible explanation is that we evaluate the model on the pairwise ranking task, whereas it is trained on the auxiliary classification task. 
Prior work has shown that fine-tuning on an auxiliary task can improve performance on a target task \citep{Pruksachatkun2020IntermediateTaskTL}. In our case, moderate amounts of classification data indeed help the model to align the embedding space in a way that benefits ranking. However, relying on larger amounts of classification data may cause the model to specialize too strongly to the auxiliary objective, reducing its effectiveness for the ranking task.

\paragraph{Impact of Overlap on Wikidata-1 Results}
In Table \ref{tabNonPerceptualDatasets}, we reported results on Wikidata-1. However, two of the properties used in the classification dataset ("long river" and "populous city"), which were generated by GPT-4o, also appeared in Wikidata-1. This overlap only affects the results of Wikidata-1 for the \textsc{ProtoSim (LLMs)} approach, as other approaches do not use the classification data. To address this, we re-evaluated Wikidata-1 after removing the overlapping properties. The accuracies for \textsc{ProtoSim (LLMs)} are as follows:
\begin{itemize}
    \item Llama3-8B: 66.3\%
    \item Mistral-24B 67.8\%
\end{itemize}
The accuracies are very similar to the values reported in Table \ref{tabNonPerceptualDatasets}, so the conclusions remain unchanged. Note that for the perceptual datasets, there is no overlap between the properties being tested and those that appear in the classification training dataset.

\begin{table}[t]
\footnotesize
\centering
\setlength\tabcolsep{2.85pt}
\begin{tabular}{lrrrrrr}
\toprule
Label & $N$ & Acc$_{\text{class}}$ & Acc$_{\text{class+rank-perc}}$ & $n_{10}$ & $n_{01}$ & $p$-value \\
\midrule
Sweetness  & 340 & 77.4 & 78.2 & 4  & 7  & 0.549 \\
Saltiness  & 340 & 78.5 & 79.4 & 5  & 8  & 0.581 \\
Sourness   & 340 & 69.7 & 70.0 & 7  & 8  & 1.000 \\
Bitterness & 340 & 65.0 & 60.3 & 23 & 7 & 0.0052 \\
Umaminess  & 340 & 70.9 & 73.2 & 3  & 11 & 0.057 \\
Fattiness  & 340 & 72.6 & 75.3 & 7  & 16 & 0.093 \\
\bottomrule
\end{tabular}
\caption{McNemar’s test results comparing classification-only vs.\ ranking+classification on the Taste dataset. $n_{10}$: correct by classification-only but not ranking+classification, $n_{01}$: correct by ranking+classification but not classification-only.}
\label{tab:mcnemar}
\end{table}

\begin{table}[t]
\footnotesize
\centering
\setlength\tabcolsep{2.9pt}
\begin{tabular}{cccccc}
\toprule
\rotatebox{90}{\textbf{Sweetness}} &  \rotatebox{90}{\textbf{Saltiness}} &  \rotatebox{90}{\textbf{Sourness}} &  \rotatebox{90}{\textbf{Bitterness}} &  \rotatebox{90}{\textbf{Umaminess}} &  \rotatebox{90} {\textbf{Fattiness}}\\
\midrule
0.75	&	0.62	&	0.56	&	0.28	&	0.52 &	0.56\\ 
\bottomrule
\end{tabular}
\caption{Pearson's correlation coefficient \textit{(r)} between the predicted scores and the ground truth human ratings, for the Taste dataset. The results were using the ProtoSim  \emph{Class + rank-perc} model.\label{tabPearsonCorrTaste}}
\end{table}

\begin{table*}[t]
\footnotesize
\centering
\setlength\tabcolsep{3pt}

\begin{tabular}{ccccccc cccc ccccccccc}
\toprule
\multicolumn{7}{c}{\textbf{Rocks}} & \multicolumn{4}{c}{\textbf{Odour}} & \multicolumn{9}{c}{\textbf{Music}}\\
\cmidrule(lr){1-7}\cmidrule(lr){8-11}\cmidrule(lr){12-20}
  \rotatebox{90}{\textbf{Lightness}} &  \rotatebox{90}{\textbf{Grain size}} &  \rotatebox{90}{\textbf{Roughness}} &  \rotatebox{90}{\textbf{Shininess}} &  \rotatebox{90}{\textbf{Organisation}}  &  \rotatebox{90}{\textbf{Variability}} &  \rotatebox{90}{\textbf{Density}} &  \rotatebox{90}{\textbf{Familiarity}} & \rotatebox{90}{\textbf{Intensity}} & \rotatebox{90}{\textbf{Pleasantness}} & \rotatebox{90}{\textbf{Irritability}}
&  \rotatebox{90}{\textbf{Wonder}} &  \rotatebox{90}{\textbf{Transcendence}} &  \rotatebox{90}{\textbf{Tenderness}} &  \rotatebox{90}{\textbf{Nostalgia}} &  \rotatebox{90}{\textbf{Peacefulness}} &  \rotatebox{90}{\textbf{Energy}} &  \rotatebox{90}{\textbf{Joyful activation}} &  \rotatebox{90}{\textbf{Sadness}} &  \rotatebox{90}{\textbf{Tension}} \\
\midrule
     0.71&   0.23&	0.01&	0.36&	0.39&	0.30&	 0.55&     0.01&	0.24&    0.49&	0.27&	    0.23&	0.48&	0.51&	0.44&	0.46&	0.32&	0.48&	0.43&	0.47\\
\bottomrule
\end{tabular}
\caption{
Pearson's correlation coefficient \textit{(r)} between the predicted scores and the ground truth human ratings, for the Rocks, Odour and Music datasets. The results were using the ProtoSim  \emph{Class + rank-perc} model.
\label{tabPearsonCorrelationDatasets}}
\end{table*}



\end{document}